\documentclass{article}

\usepackage[preprint]{neurips_2026}

\usepackage[utf8]{inputenc} 
\usepackage[T1]{fontenc}    
\usepackage{hyperref}       
\usepackage{url}            
\usepackage{booktabs}       
\usepackage{amsfonts}       
\usepackage{nicefrac}       
\usepackage{microtype}      
\usepackage{xcolor}         
\usepackage{booktabs}
\usepackage{colortbl}
\usepackage{graphicx}
\usepackage{xcolor}
\usepackage{float}
\usepackage{amsmath}
\usepackage{wrapfig}
\usepackage[capitalize,noabbrev]{cleveref}
\usepackage{xspace}
\usepackage{xcolor} 
\usepackage{multirow}
\usepackage{graphicx} 
\newcommand{\pos}[1]{{\scriptsize\textcolor{green!60!black}{+#1}}}
\newcommand{\dec}[1]{{\scriptsize\textcolor{red!70!black}{-#1}}}
\usepackage{lineno}

\definecolor{darkblue}{rgb}{0, 0, 0.5}
\hypersetup{colorlinks=true, citecolor=darkblue, linkcolor=darkblue, urlcolor=darkblue}

\makeatletter
\renewcommand{\paragraph}{%
  \@startsection{paragraph}{4}{\z@}%
                {0.6ex \@plus 0.3ex \@minus 0.1ex}%
                {-1em}%
                {\normalsize\bf}%
}
\makeatother

\newcommand{\ours}{\textbf{\textsc{\textcolor[HTML]{1a56e8}{W}\textcolor[HTML]{3b44e2}{e}\textcolor[HTML]{5c32db}{b}\textcolor[HTML]{7e22ce}{G}\textcolor[HTML]{941cbb}{r}\textcolor[HTML]{aa16a7}{a}\textcolor[HTML]{bf1192}{p}\textcolor[HTML]{cc1077}{h}\textcolor[HTML]{d41059}{M}\textcolor[HTML]{d91243}{i}\textcolor[HTML]{dc1832}{x}}}\xspace}
\newcommand{\oursplus}{\textbf{\textsc{\textcolor[HTML]{1a56e8}{W}\textcolor[HTML]{3b44e2}{e}\textcolor[HTML]{5c32db}{b}\textcolor[HTML]{7e22ce}{G}\textcolor[HTML]{941cbb}{r}\textcolor[HTML]{aa16a7}{a}\textcolor[HTML]{bf1192}{p}\textcolor[HTML]{cc1077}{h}\textcolor[HTML]{d41059}{M}\textcolor[HTML]{d91243}{i}\textcolor[HTML]{dc1832}{x}\textcolor[HTML]{dc2626}{+}}}\xspace}
\newcommand{\topk}{Top-$K$\xspace}
\newcommand{\bottomk}{Bottom-$K$\xspace}
\newcommand{\opaddsub}{\texttt{+-}\xspace}
\newcommand{\opmuldiv}{\texttt{*/}\xspace}

\newcommand{\iconfallback}[1]{\raisebox{0.1pt}{\scriptsize\texttt{#1}}\xspace}
\newcommand{\projecticon}{%
  \IfFileExists{img/logos/project_page_icon.png}
    {\raisebox{-1.2pt}{\includegraphics[height=1.05em]{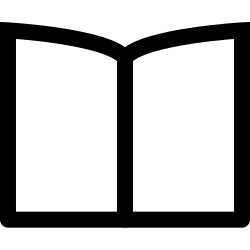}}\xspace}
    {\iconfallback{BK}}%
}
\newcommand{\github}{%
  \IfFileExists{img/logos/github_logo.pdf}
    {\raisebox{-1.3pt}{\includegraphics[height=1.05em]{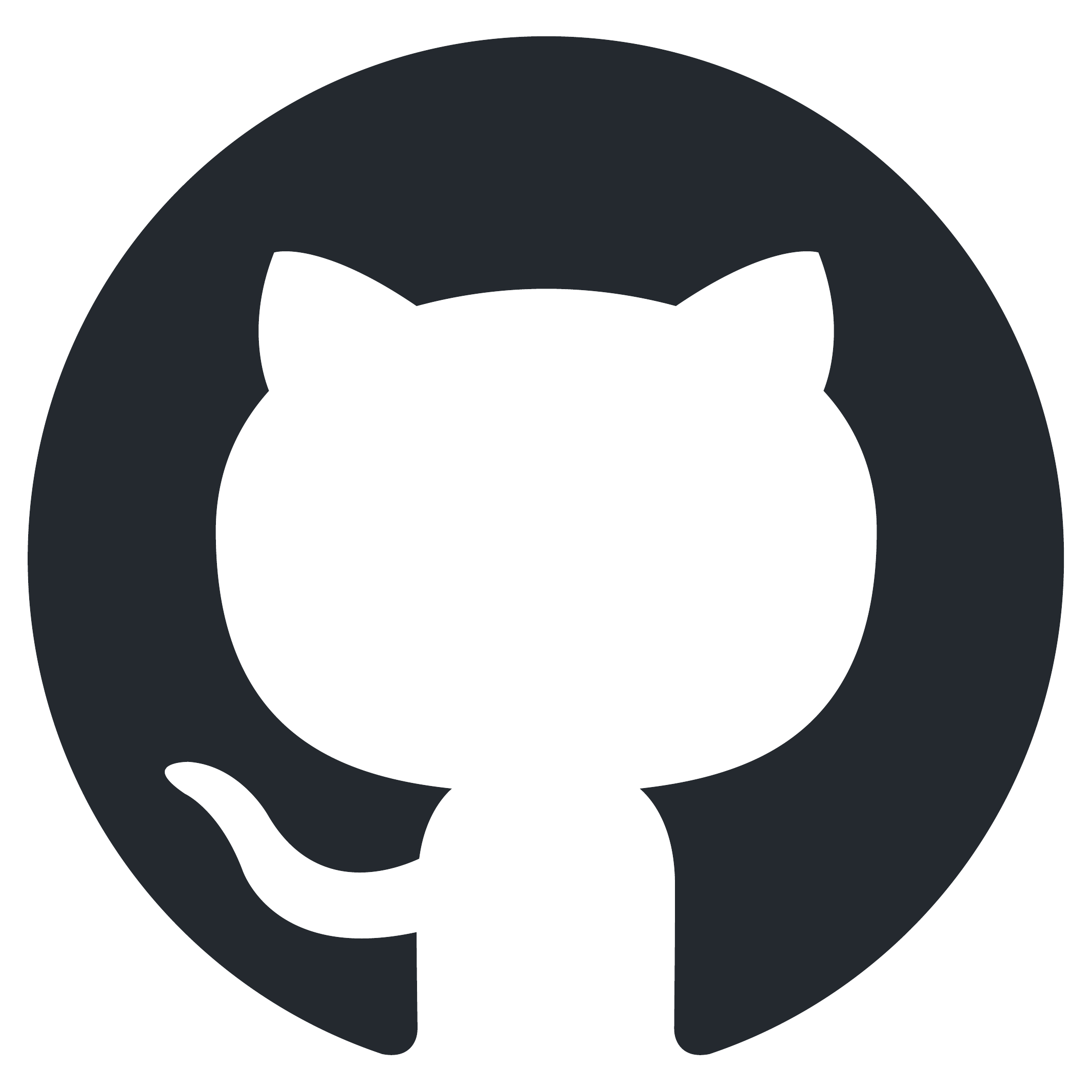}}\xspace}
    {\iconfallback{GH}}%
}
\newcommand{\hf}{%
  \IfFileExists{img/logos/huggingface_logo.pdf}
    {\raisebox{-1.3pt}{\includegraphics[height=1.05em]{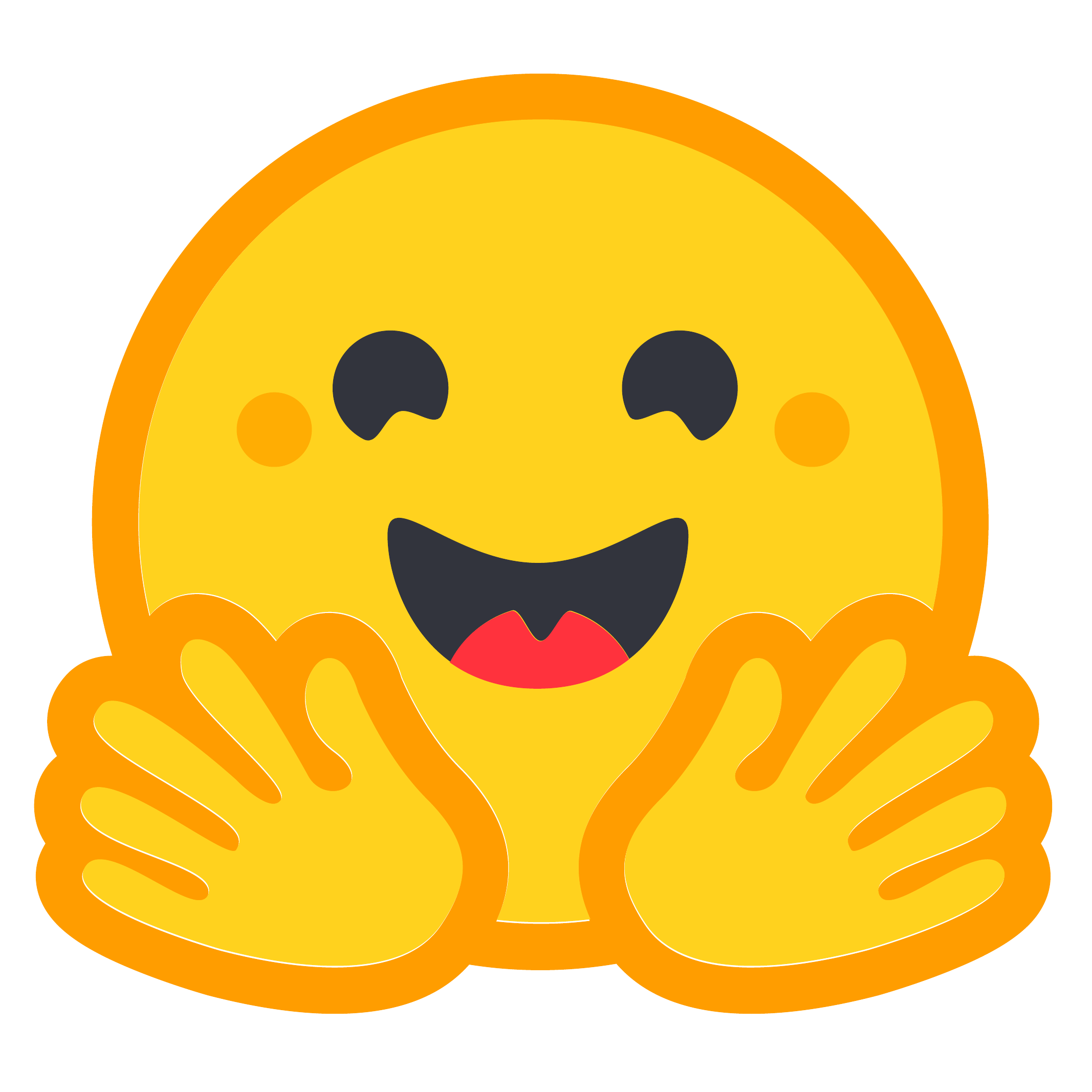}}\xspace}
    {\iconfallback{HF}}%
}
\newcommand{\dbicon}{%
  \IfFileExists{img/logos/data_icon.pdf}
    {\raisebox{-1.3pt}{\includegraphics[height=1.05em]{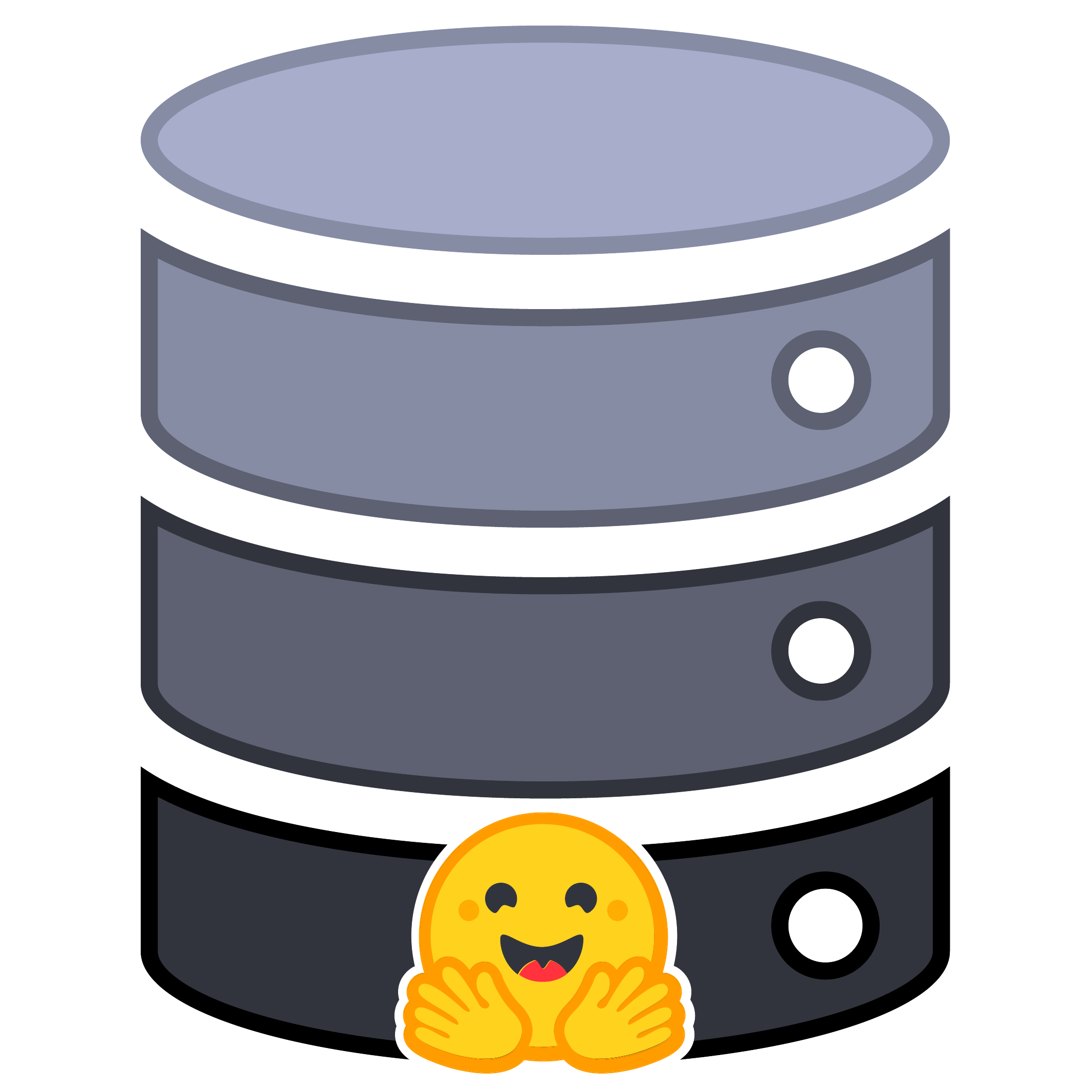}}\xspace}
    {\iconfallback{DB}}%
}
\newcommand{\worldwideweb}{\projecticon}

\newcommand{\rescourcelink}{%
    \begin{center}
        \small
        \worldwideweb\,\href{https://princeton-pli.github.io/WebGraphMix/}{\textbf{Project Page}}
        \qquad
        \github\,\href{https://github.com/princeton-pli/WebGraphMix}{\textbf{Code}}
        \qquad
        \hf\,\href{https://huggingface.co/collections/PrincetonPLI/webgraphmix}{\textbf{Hugging Face}}
    \end{center}%
}

\title{Hubs or Fringes: Pretraining Data Selection \\ via Web Graph Centrality}

\author{%
  Vedant Badoni \qquad Danqi Chen \qquad Xinyi Wang\\
  \texttt{\{vedantbadoni, danqic\}@princeton.edu} \quad \texttt{wangxinyilinda@gmail.com} \\
  Princeton Language and Intelligence
}

\begin{document}

\maketitle

\vspace{-0.85cm}
\rescourcelink
\vspace{0.2cm}




\begin{abstract}

The performance of modern language models depends critically on pretraining data composition. Yet existing data selection methods rely on auxiliary classifiers for document scoring or mixture optimization, adding computational overhead and dependence on labeled data. We propose \ours, a lightweight
data selection framework that computes structural centrality scores over the Common Crawl host-level web graph and uses them to vary the proportion of central versus peripheral documents in the pretraining mixture. We hypothesize that central hosts expose models to reusable abstractions, while peripheral hosts encode specialized, long-tail knowledge.
\ours computes centrality scores efficiently at web scale, requiring no model training, labeled data, or downstream supervision.
We integrate \ours into the DataComp-LM pipeline and train models at 400M and 1B parameter scales with 8B and 28B tokens respectively, evaluating on 23 tasks ranging from factual knowledge to symbolic reasoning.
Our experiments show that central and peripheral web regions encode complementary capabilities. Mixture combining both at a ratio of 1:1 achieves 41.4\% on average, compared to 39.8\% for uniform sampling. Combining structural scores with document-level quality classifier scores further improves performance to 43.8\%. These findings demonstrate that web graph topology is a meaningful axis for pretraining data curation, capturing information that is largely orthogonal to existing~content-based~approaches.

\end{abstract}

\section{Introduction}

The performance of modern language models (LMs) depends critically on the composition of their pretraining data.
While neural scaling laws~\citep{kaplan2020scaling, hoffmann2022training} characterize how data size affects performance, far less is understood about how the structure of large-scale web corpora should influence data selection.
In practice, modern pretraining pipelines rely on massive web dumps that are filtered, deduplicated, and sampled at the document level~\citep{albalak2024survey}. These pipelines implicitly treat documents as independent units, applying heuristic quality filters or domain classifiers without considering relationships between documents~\citep{soldaini2024dolma}. As a result, existing approaches largely ignore how information is organized across the web.

However, the web is fundamentally a graph. Webpages and hosts are connected through hyperlinks, forming a large-scale network that encodes topical structure, citation patterns, and information flow. We hypothesize that a document’s structural position in this graph may correlate with the type and transferability of knowledge it provides during pretraining. Structurally central documents—those that lie on many shortest paths or connect diverse regions—act as hubs or bridges between otherwise weakly connected communities, and are more likely to co-occur with heterogeneous contexts and expose models to reusable abstractions. In contrast, peripheral documents may encode specialized or long-tail content that is less broadly shared. From a language modeling perspective, this suggests that graph structure may influence the diversity and overlap of token-level learning signals, and therefore shape the capabilities learned during pretraining.

\begin{wrapfigure}{r}{0.5\textwidth}
  \begin{center}
    \includegraphics[width=0.48\textwidth, trim=0 0 0 16pt, clip]{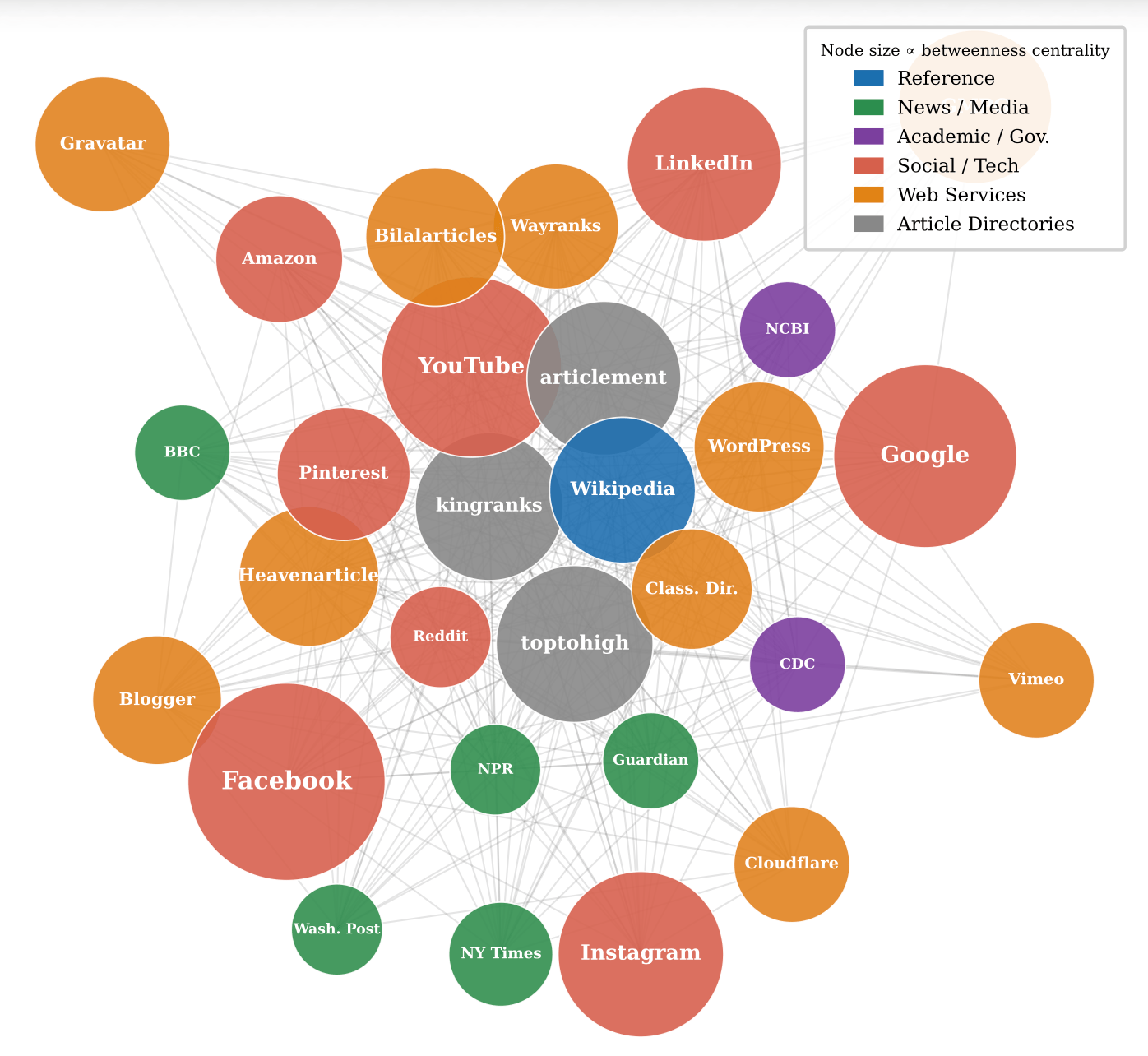}
  \end{center}
   \caption{Subgraph of the Common Crawl host-level web graph. Node size is proportional to their Betweenness centrality score.}
  \label{fig:graph}
\end{wrapfigure}

In this work, we introduce {\ours}, a graph-based data selection framework that leverages web-scale structural signals to construct pretraining mixtures. 
\ours operates directly on the hyperlink graph and is fully unsupervised. We compute centrality measures over a large Common Crawl host-level graph and use these scores to partition data into structurally distinct subsets. We then construct training mixtures that emphasize (i) structurally \textit{central} data, (ii) structurally \textit{peripheral} data, and (iii) combinations of the two, enabling controlled investigation of how graph position affects downstream model behavior. We test mainly two ways of computing graph centrality: Betweenness centrality~\citep{freeman1977betweenness} and Katz centrality~\citep{katz1953}. We also tried PageRank-based~\citep{pagerank1999} scoring but failed to show improvement, which is consistent with the observation from DCLM~\mbox{\citep{li2024datacomp}}.

\ours differs from the prior domain-based and quality-based approaches. Domain-based methods construct semantic taxonomies (e.g., topic and format categories)~\citep{wettig2025weborganizer} or optimize coarse-grained domain mixtures (e.g., arXiv, GitHub, Common Crawl) through regression or proxy training~\citep{xie2023doremi, liu2025regmix}, while quality-based methods score documents by abstract qualities (e.g., educational value, difference between raw web and curated high-quality data)~\citep{penedo2024fineweb, sachdeva2026how, wettig2024qurating, li2024datacomp, gunasekar2023textbooks}.
In contrast, \ours does not require a taxonomy, classifier, or regression model—only structural signals intrinsic to the web graph—making it lightweight and directly transferable across corpora that expose hyperlink structure.

We integrate \ours into the standardized DataComp-LM (DCLM) pipeline~\citep{li2024datacomp} and train models at 400M and 1B parameter scales with 8B and 28B tokens, respectively. Centrality scores for the full Common Crawl host graph (13.9M nodes, 439.6M edges) take fewer than 9 GPU hours to compute in total and can then be reused across all downstream experiments. All training runs use identical tokenization, shuffling, and optimization procedures to isolate the effect of data selection, and we evaluate on a wide range of 23 tasks from the DCLM CORE v2 benchmark~\mbox{\citep{li2024datacomp}}.

Our results show that graph structure provides a meaningful and complementary signal for pretraining data curation. At 1B scale, selecting documents from structurally central hosts
improves performance on Symbolic \& Algorithmic Reasoning by +1.4\% over uniform sampling, while selecting from peripheral hosts improves Science \& Factual Knowledge and Commonsense \& Reasoning. These opposing effects indicate that different regions of the web graph encode distinct capability-relevant signals, and motivate mixture sampling: combining 50\% central and 50\% peripheral documents with betweenness centrality reaches 41.4\% average across all 23 tasks, compared to 39.8\% for uniform sampling. Combining the centrality signal with the DCLM-fasttext quality classifier through multiplicative \& divisive scoring further improves performance to 43.8\%, indicating that web graph topology captures information that is largely orthogonal to content-based quality signals.


Together, our results suggest that treating the web as a structured graph—rather than an unordered corpus—opens a new direction for studying the relationship between data distribution and \mbox{model capabilities}.

\section{Related Work}


\paragraph{Heuristic filtering \& deduplication.} Existing approaches to data curation largely operate at the document level and treat documents as independent units. 
The first stage of curation usually applies heuristic filtering and deduplication. Rule-based filters remove boilerplate, spam, and malformed text~\citep{raffel2020exploring, rae2021scaling, penedo2023the}, while deduplication techniques such as MinHash~\citep{Broder1997OnTR, lee-etal-2022-deduplicating} and Bloom-filter-based methods~\citep{soldaini2024dolma} eliminate near-duplicate documents to reduce memorization. Frameworks such as DataComp-LM (DCLM)~\citep{li2024datacomp} standardize these preprocessing steps and enable compute-controlled comparisons. While effective at improving data cleanliness and diversity, these methods do not model relationships between documents.

\paragraph{Document quality scoring.} The second stage of curation usually assigns scalar quality scores to documents and selects data based on ranking. FineWeb-Edu~\citep{penedo2024fineweb}, DCLM-fasttext~\citep{li2024datacomp}, QuRating~\citep{wettig2024qurating}, and Ask-LLM~\citep{sachdeva2026how}
estimate properties such as educational value or difference between curated high-quality corpora and low-quality corpora.
Benchmark-Targeted Ranking (BETR)~\citep{mizrahi2025language} explicitly aligns pretraining data with downstream tasks by selecting documents similar to benchmark examples, achieving substantial gains under scaling-law analysis. Other approaches use perplexity~\citep{wenzek-etal-2020-ccnet}, n-gram overlap~\citep{xie2023doremi}, or attention-based signals~\citep{hua2025attentioninfluence} to identify useful data. Despite their diversity, these methods share a common formulation: data selection is treated as a ranking problem over independently scored documents.

\paragraph{Domain mixture optimization.} The third stage of curation usually introduces higher-level structure by partitioning web data into domains and optimizing mixture weights. Most of the work like DoReMi~\citep{xie2023doremi}, RegMix~\citep{liu2025regmix}, TiKMiX~\citep{wang2025tikmix}, DoGE~\citep{fan2024doge}, and Aioli~\citep{chen2025aioli} use a coarse-grained, pre-defined domain categorization and optimize over the weights of mixtures using proxy models, regression, or influence-based techniques.
To demystify the domain taxonomy of pretraining data, work like Skill-it~\citep{chen2023skillit}, WebOrganizer~\citep{wettig2025weborganizer}, Nemotron-CLIMB~\citep{diao2026nemotronclimb}, and Group-MATES~\citep{yu2026grouplevel} defines their own data domains before optimizing the mixture, by either clustering or constructing a compact and interpretable domain taxonomy.
These approaches can yield strong empirical gains, but typically require substantial computation, model training, \mbox{or downstream supervision}.

Underlying all these approaches is a shared assumption: documents are evaluated primarily based on their content or similarity, rather than on how they relate to one another. Even when structure is introduced (e.g., domains or clusters), it is derived from semantic similarity or learned representations, not from the native connectivity of the web.

\paragraph{Useful web graph structure.}
In contrast, The web is fundamentally a graph: hyperlinks connect pages and hosts into a large-scale network encoding citation, topical proximity, and information flow. Graph-based methods such as PageRank~\citep{pagerank1999} and HITS~\citep{kleinberg1999hits} have long exploited this structure for ranking and retrieval. Recent work, Craw4LLM~\citep{yu-etal-2025-craw4llm}, introduces quality-aware crawling to improve crawler efficiency---using webpage quality as the crawler scheduler's priority score rather than graph connectivity, reducing crawled pages to 21\% of the baseline while matching its performance.
While Craw4LLM incorporates quality signals during crawling, we reintroduce web graph structure \emph{after} crawling for data selection.
A complementary direction leverages web metadata at training time: MeCo~\citep{gao2025metadata} conditions on URL information to improve data efficiency and enable controllable inference, with gains persisting even under URL anonymization---suggesting that grouping documents by source provides useful structural signal. Unlike these approaches, our method operates purely at the data selection stage.

To the best of our knowledge, prior work has not used graph-theoretic position as a direct signal for selecting and weighting documents within an already-crawled corpus for pretraining.



\section{Our Method: \ours}

We introduce {\ours}, a lightweight pretraining data selection framework that leverages structural signals from the web graph. Rather than scoring documents independently based on content, our method assigns \emph{centrality scores} based on each document’s position in the global hyperlink network and uses these scores to guide sampling.

\subsection{Web Graph Construction}

We operate on the Common Crawl host-level graph\footnote{We use \texttt{cc-main-2023-24-sep-nov-feb-host} from \url{https://commoncrawl.org/web-graphs}.}, where each node represents a web host (e.g., \texttt{wikipedia.org}) and directed edges correspond to hyperlinks between hosts. Formally, we define a directed graph $G = (V, E)$, where $v \in V$ denotes a host and $(u, v) \in E$ indicates that host $u$ links to host $v$. This host-level representation aggregates all documents from the same domain into a single node, yielding a large-scale graph with 13.9M nodes and 439.6M edges. 

The raw pretraining corpus we use, Corpus-200B\footnote{\url{https://huggingface.co/datasets/WebOrganizer/Corpus-200B}} from \cite{wettig2025weborganizer}, is a pre-processed version of the \texttt{1b-1x} CommonCrawl pool from DataComps-LM~\citep{li2024datacomp} cleaned with RefinedWeb filters~\citep{penedo2023the} and BFF deduplication~\citep{groeneveld2024bff}. Each document in the preprocessed corpus is mapped to its corresponding host via its URL. We discard about 5\% of the documents in the corpus without a host in the web graph. Centrality scores are computed at the host level and inherited by all associated documents. Specifically, if a host $v$ has centrality score $c(v)$, then each document $d_i$ from that host is assigned score $s_i = c(v)$.

\subsection{Centrality Score}

We quantify structural importance using classical graph centrality measures that capture complementary aspects of connectivity.

\textbf{Betweenness centrality}~\citep{freeman1977betweenness} measures how frequently a node lies on shortest paths between other nodes:
\begin{equation}
c_B(v) = \sum_{s \neq v \neq t} \frac{\sigma(s,t \mid v)}{\sigma(s,t)},
\end{equation}
where $s, t, v \in E$, $\sigma(s,t)$ is the number of shortest paths from node $s$ to node $t$, and $\sigma(s,t \mid v)$ counts those passing through node $v$. Nodes with high betweenness act as bridges between otherwise weakly connected regions. Representated crawled content from the hosts with the highest and lowest betweenness centrality scores are shown in \cref{tab:centrality_content_contrast}.

\begin{table}[t]
\centering
\caption{
\textbf{Host centrality reflects different types of web content.}
High-betweenness hosts tend to contain broadly reusable, cross-domain patterns, whereas low-betweenness hosts more often contain specialized or long-tail information.
Examples are based on representative URLs observed in the crawl. For actually text crawled from these URLs, see Appendix~\ref{app:snippet}.
}
\label{tab:centrality_content_contrast}
\footnotesize
\setlength{\tabcolsep}{4pt}
\renewcommand{\arraystretch}{1.1}

\definecolor{HighBlueDark}{HTML}{1F5AA6}
\definecolor{LowRedDark}{HTML}{B03A2E}

\begin{tabular}{
    >{\raggedright\arraybackslash}p{2.7cm}
    >{\raggedright\arraybackslash}p{5.8cm}
    >{\raggedright\arraybackslash}p{4.4cm}
}
\toprule
\textbf{Host (score)}
& \textbf{Representative crawled content}
& \textbf{Knowledge tendency} \\
\midrule

\multicolumn{3}{l}{\cellcolor{blue!12}\textbf{\textcolor{HighBlueDark}{High betweenness / central hosts}}} \\

\cellcolor{blue!20} \texttt{facebook.com} {\scriptsize($1.98{\times}10^{-1}$)}
& \cellcolor{blue!20} Public social/profile/community content, including posts, pages, groups, photos, and videos.
& \cellcolor{blue!20} Broad social and common-sense language patterns. \\

\cellcolor{blue!16} \texttt{google.com} {\scriptsize($1.18{\times}10^{-1}$)}
& \cellcolor{blue!16} Aggregated search- and finance-style information, including market summaries, entities, \mbox{and linked news}.
& \cellcolor{blue!16} Cross-domain reference and retrieval-style knowledge. \\

\cellcolor{blue!12} \texttt{youtube.com} {\scriptsize($1.07{\times}10^{-1}$)}
& \cellcolor{blue!12} Video pages, titles, descriptions, transcripts, comments, and media metadata.
& \cellcolor{blue!12} Instructional, cultural, and explanatory content. \\

\cellcolor{blue!9} \texttt{linkedin.com} {\scriptsize($3.64{\times}10^{-2}$)}
& \cellcolor{blue!9} Professional profiles, companies, roles, skills, recommendations, and work histories.
& \cellcolor{blue!9} Occupational and organizational knowledge. \\

\cellcolor{blue!6} \texttt{en.wikipedia.org} {\scriptsize($2.46{\times}10^{-2}$)}
& \cellcolor{blue!6} Encyclopedia-style articles covering scientific, historical, cultural, and technical topics.
& \cellcolor{blue!6} High-coverage factual and conceptual knowledge. \\

\midrule

\multicolumn{3}{l}{\cellcolor{red!12}\textbf{\textcolor{LowRedDark}{Low betweenness / peripheral hosts}}} \\

\cellcolor{red!5} \texttt{hammarsdrama.com} {\scriptsize($4.41{\times}10^{-20}$)}
& \cellcolor{red!5} Performing-arts and dance-film production company information.
& \cellcolor{red!5} Narrow institutional and arts-domain knowledge. \\

\cellcolor{red!7} \texttt{ontoma.com} {\scriptsize($3.86{\times}10^{-20}$)}
& \cellcolor{red!7} FinTech platform responding to a specific financial-services regulatory context.
& \cellcolor{red!7} Specialized industry and regulatory information. \\

\cellcolor{red!9} \texttt{bluepenstrokes.com} {\scriptsize($3.28{\times}10^{-20}$)}
& \cellcolor{red!9} Personal creative writing, daily reflections, and artistic prose.
& \cellcolor{red!9} Individual voice and long-tail expressive content. \\

\cellcolor{red!11} \texttt{abcsofsex-ed.org} {\scriptsize($1.98{\times}10^{-20}$)}
& \cellcolor{red!11} Workshop material for a specific sex-education and public-health setting.
& \cellcolor{red!11} Specialized educational and local-program knowledge. \\

\cellcolor{red!14} \texttt{riomardesigns.com} {\scriptsize($6.82{\times}10^{-21}$)}
& \cellcolor{red!14} Small-site travel-health advice and informational content.
& \cellcolor{red!14} Niche advice and domain-specific long-tail information. \\

\bottomrule
\end{tabular}
\end{table}

\textbf{Katz centrality}~\citep{katz1953} captures recursive influence by aggregating contributions from all walks in the graph:
\begin{equation}
c_K(v_i) = \eta \sum_j A_{ij} c_K(v_j) + \tau,
\end{equation}
where $A$ is the adjacency matrix, $v_i, v_j \in E$, $i$ and $j$ both index all nodes in the graph, $0 < \eta < 1 / \lambda_{\max}$ ensures convergence, and $\tau$ is a bias term. This assigns higher scores to nodes connected to other influential nodes, while attenuating longer paths. These measures capture complementary notions of structural importance: Betweenness emphasizes cross-community connectivity, while Katz centrality reflects global influence.

\textbf{PageRank}~\citep{pagerank1999} is a specific variant of eigenvector centrality. Prior work has shown that eigenvector centrality can be used in place of PageRank in directed networks with lower computational cost while preserving rank correlation~\citep{chandrashekhar2022pagerank}. We ran ablations using eigenvector centrality but found it did not yield improvements over the baseline. A similar conclusion was reached by DCLM~\citep{li2024datacomp}: they find varying the top quantile data selection based on PageRank scores does not outperform uniform sampling. Thus we focus on Betweenness and Katz centrality in the main paper instead as they are shown to be effective and capture distinct and complementary notions of structural importance---bridging versus weighted influence. 

\paragraph{Efficiency and scalability.}
A key advantage of \ours is that centrality scores can be computed efficiently at web scale using distributed graph algorithms. We implement centrality computation over the host graph using GPU-parallelized primitives and graph partitioning with the \texttt{cuGraph} library\footnote{\url{https://github.com/rapidsai/cugraph}}. In practice, computing Katz centrality~\citep{Foster2001} for the full Common Crawl host graph took us < 3 hours on one H100 GPU and computing Betweenness centrality~\citep{brandes2001betweenness} took us < 6 hours on 4 H100 GPUs, after which the scores can be reused across all~\mbox{downstream experiments}.

Unlike prior data selection methods that require repeated model training, gradient computation, or proxy evaluation, this is a \emph{compute-efficient one-time preprocessing step}.
Once computed, centrality scores incur negligible overhead during data sampling.

\subsection{Centrality-Guided Data Sampling}

Each host can be viewed as a subdomain embedded within the global web graph. Hosts differ substantially in their structural roles: some connect multiple regions of the graph and act as hubs or bridges, while others lie in sparsely connected or peripheral regions.  We hypothesize that these structural differences correspond to differences in the type of knowledge encoded: Structurally central hosts are more likely to expose models to broadly reusable and cross-domain patterns, whereas peripheral hosts tend to contain specialized or long-tail information. This can qualitatively observed from \cref{tab:centrality_content_contrast}, where we show crawled content of central hosts and peripheral hosts.
To study this effect, we construct data mixtures that vary systematically across the centrality spectrum.

Given host-level centrality scores $c(v)$, each document inherits a score $s_i = c(v_i)$ based on its host $v_i$. We then construct training datasets under a fixed token budget using the \mbox{following sampling strategies}.

\textbf{\topk (Central) sampling}:
We select documents whose hosts fall within the top percentile of the centrality distribution (e.g., top 25\%, or 75\%), emphasizing structurally central regions of the web.

\textbf{\bottomk (Peripheral) sampling}:
We select documents from the lowest percentile of the centrality distribution, focusing on peripheral or long-tail regions.

\textbf{Mixed sampling}:
To test whether central and peripheral regions provide complementary signals, we construct mixtures combining both strata:
\begin{align}
    \label{eq:mixture}
    \alpha \cdot \text{\topk} + (1-\alpha) \cdot \text{\bottomk},
\end{align}

where the mixture ratio $\alpha \in \{0, 0.25, 0.5, 0.75, 1\}$. Documents are sampled proportionally until the token budget is reached. 

\subsection{Combining Structural and Quality Signals}

In addition to pure structural selection, we explore combining centrality scores with document-level quality scores. We use the quality scores produced by DCLM-fasttext~\citep{li2024datacomp}, a bigram model trained to classify high quality text sampled from different sources and low quality text sampled from RefinedWeb~\citep{penedo2023the} reproduction.
We normalize both the centrality scores and the quality scores by:
\begin{align}
    \hat{s}_i = \exp(s_i - \max_j s_j),
\end{align}
where $i$ and $j$ both index all hosts in the web graph. This gives us a score within $(0, 1]$. 
After normalizing both signals, we combine graph centrality and document quality in two complementary ways. For \topk selection, we use additive and multiplicative scores,
\[
\hat{s}_i^{\mathrm{add}}=\hat{s}_i^{\mathrm{centrality}}+\hat{s}_i^{\mathrm{quality}},
\qquad
\hat{s}_i^{\mathrm{mult}}=\hat{s}_i^{\mathrm{centrality}}\cdot \hat{s}_i^{\mathrm{quality}},
\]
which favor documents that are both central in the web graph and high quality. For \bottomk selection, we instead use contrastive scores,
\[
\hat{s}_i^{\mathrm{sub}}=\hat{s}_i^{\mathrm{centrality}}-\hat{s}_i^{\mathrm{quality}},
\qquad
\hat{s}_i^{\mathrm{div}}=\hat{s}_i^{\mathrm{centrality}}/\hat{s}_i^{\mathrm{quality}},
\]
and select documents with the lowest scores, thereby prioritizing high-quality documents that are less central. Documents are ranked by the corresponding combined score and selected under the same token budget. These strategies allow us to test whether graph structure provides a signal complementary to document quality.

\section{Experiments}

\subsection{Experimental Setup}

All experiments are conducted using the official DataComp‑LM (DCLM) framework, which provides standardized data pools, fixed model architectures, compute-optimal token budgets, and a fully reproducible training and evaluation pipeline. 
We evaluate two compute scales: \texttt{400m‑1x}, which trains a 412M-parameter model on approximately 8.2B tokens, and \texttt{1b‑1x}, which trains a 1.4B-parameter model on approximately 28B tokens. We mainly report 1B model results in the main paper as they are more significant. Full 400M model results can be found in Appendix~\ref{app:results}. 

We report task-level and average accuracy on DCLM CORE v2 benchmark~\citep{li2024datacomp}, which consists of 23 tasks. As described in \cref{tab:eval-tasks} in Appendix~\ref{app:exp}, the eval tasks are classified into 5 categories\footnote{As marked in meta data: \url{https://github.com/mlfoundations/dclm/blob/main/eval/eval_meta_data.csv}}: Commonsense \& Reasoning, QA \& Comprehension, Science \& Factual Knowledge, Symbolic \& Algo Reasoning, and Language Understanding. In the following paper, we will use \emph{Commonsense}, \emph{Comprehension}, \emph{Knowledge}, \emph{Reasoning}, and \emph{Language} as abbreviations.
We also look into each category of tasks to get a better understanding of the distinct effect of \topk (central) and \bottomk (peripheral) sampling.


We consider the following baselines: \textbf{Random}: uniformly randomly select from the data pool; \textbf{Quality}: select documents with top K quality score produced by DCLM-fasttext; \textbf{WebOrganizer}~\citep{wettig2025weborganizer}: topic and format domain pairs mixture predicted from RegMix~\citep{liu2025regmix} pipelines; \textbf{WebOrganizer+}~\citep{wettig2025weborganizer}: the WebOrganizer domain mixture combined with DCLM-fasttext quality filter; \textbf{PageRank}: select documents with top K eigenvector centrality which can be used in place of the classical PageRank algorithm~\citep{chandrashekhar2022pagerank}.

\subsection{Main Results}
\label{sec:4.1}



Recall that \ours selects pretraining data by computing host-level centrality scores over the Common Crawl web graph and constructing a \emph{mixed} dataset that combines documents from the top (central) and bottom (peripheral) quantiles of the centrality distribution, with mixture ratio $\alpha$ controlling the balance between the two strata (see \cref{eq:mixture}).
Among the two centrality measures explored, we find \emph{betweenness centrality} most effective; it identifies hosts that bridge otherwise weakly connected regions of the web, yielding documents with broadly reusable, cross-domain patterns.
\oursplus further incorporates document-level quality scores from DCLM-fasttext: for the \topk stratum, documents are ranked by the \emph{multiplicative} combination of centrality and quality scores; for the \bottomk stratum, documents are ranked by the \emph{division} of centrality by quality, thereby surfacing high-quality documents that are structurally peripheral.

\cref{tab:1b_main_category} shows the overall benchmark performance comparison between our best methods and baselines, averaged over task categories. When mixing \topk and \bottomk documents with $\alpha=0.5$
ranked by Betweenness centrality score, our method improves upon the random selection baseline by 1.6\% on average and improves upon the \topk quality score selection baseline by 1.5\% on average. Note that using quality score alone would improve upon the random selection baseline by 2.5\%, so our method combining with quality score improves upon random selection baseline by 4\% in total. \ours improves over the random selection baseline on all task categories, while \oursplus improves over the Quality baseline on 4 out of 5 categories. 

The WebOrganizer baseline requires significant human effort for proposing the domain taxonomies. It also substantial computation: training 512 proxy models of 50M parameters and fitting a gradient-boosted regression model to optimize domain weights toward specific target tasks, namely MMLU~\citep{hendrycks2021measuring} and HellaSwag~\citep{zellers2019hellaswag}. This explains WebOrganizer+'s strong performance on Commonsense---HellaSwag is a commonsense sentence-completion benchmark---but limits the method's generalizability to other capability categories. \oursplus slightly outperforms WebOrganizer+ on overall average while requiring no proxy training, no labeled targets, and no benchmark-specific tuning.

\begin{table}[t]
\centering
\caption{Accuracy on DCLM CORE v2 benchmark at 1B scale, averaged by task category. {\ours} uses betweenness centrality with $\alpha=0.5$ \topk/\bottomk mixture; {\oursplus} additionally combines centrality with the DCLM-fasttext quality score via multiplication and division. Note that while our {\ours} is close to WebOrganizer baseline, our method is significantly cheaper and more transferable. Per-task results are reported in Appendix~\ref{app:results}.
}
\label{tab:1b_main_category}
\small
\setlength{\tabcolsep}{5pt}
\begin{tabular}{lcccccc}
\toprule
\textbf{Method} & \textbf{Commons.} & \textbf{Compreh.} & \textbf{Knowl.} & \textbf{Reason.} & \textbf{Lang.} & \textbf{Average} \\
\midrule
Random    & 57.3 & 37.9 & 34.2 & 19.0 & 39.9 & 39.8 \\
Quality   & 59.8 & 38.1 & 38.9 & 20.7 & \textbf{42.8} & 42.3 \\
WebOrganizer    & 59.6 & 39.2 & 38.0 & 22.5 & 38.3 & 42.1 \\
WebOrganizer+   & \textbf{61.9} & 41.4 & 39.1 & 21.9 & 38.8 & 43.4 \\
PageRank  & 56.9 & 37.4 & 34.8 & 19.3 & 38.1 & 39.6 \\
\ours     & 59.5 & 39.4 & 35.4 & 21.4 & 40.2 & 41.4 \\
\oursplus    & 60.8 & \textbf{42.6} & \textbf{39.7} & \textbf{22.6} & 41.9 & \textbf{43.8} \\
\bottomrule
\end{tabular}
\end{table}



The effectiveness of \ours scales with model size. As shown in \cref{tab:scaling}, the gain from the best mixture strategy over baseline grows from 0.1\% at 400M parameters to 1.6\% at 1B parameters, and the gain from combining quality scores grows from 0.6\% at 400M parameters to 1.5\% at 1B parameters. This is consistent with the scaling behavior observed in other data selection work~\citep{mizrahi2025language, yu2026grouplevel} and suggests that our method may provide larger gains at~larger~scales.
\begin{table}[t]
\centering
\caption{Best average accuracy by strategy category at 400M and 1B scale. Gain is computed relative to either the Random baseline or the Quality baseline. \opaddsub means combining with addition for \topk and subtraction for \bottomk. \opmuldiv means combining with multiplication for \topk and division for \bottomk. Full per-task results are in Appendix~\ref{app:results}.}
\label{tab:scaling}
\small
\begin{tabular}{llll}
\toprule
\textbf{Strategy} & \textbf{Best variant} & \textbf{400M} & \textbf{1B} \\
\midrule
Random & — & 32.5 & 39.8 \\
Best pure centrality & Katz Bottom-$K$ & 32.5 \pos{0.0} & 40.5 \pos{0.7} \\
Best mixture & Katz $\alpha=0.25$ / Betw. $\alpha=0.5$ & 32.6 \pos{0.1} & 41.4 \pos{1.6} \\
\midrule
Quality & — & 34.5 & 42.3 \\
Best additive & Katz (\opaddsub) $\alpha=1$ / Betw. (\opaddsub) $\alpha=0.25$ & 35.1 \pos{0.6} & 43.7 \pos{1.4} \\
Best multiplicative & Betw. (\opmuldiv) $\alpha=1$ / Betw. (\opmuldiv) $\alpha=0.5$ & 34.5 \pos{0.0} & 43.8 \pos{1.5} \\
\bottomrule
\end{tabular}
\end{table}

\section{Analysis}

\subsection{Structural Position Differentially Affects Capability Categories}


\cref{tab:knowledge_vs_reasoning_1b} breaks down performance by capability category for \topk (central) and \bottomk (peripheral) sampling strategies at the 1B scale. The results show that the effect of structural position is highly capability-dependent, and that central and peripheral regions of the web encode different types of useful information.

\paragraph{\bottomk sampling consistently improves factual and commonsense knowledge.}
The clearest and most consistent pattern appears in \emph{Knowledge} and \emph{Commonsense} task categories. In \emph{Knowledge}, both \bottomk strategies outperform the random baseline: Betweenness \bottomk improves from 34.2\% to 35.4\% (\pos{1.2\%}), while Katz \bottomk reaches 35.3\% (\pos{1.1\%}). In contrast, Betweenness \topk slightly hurts performance (\dec{0.3\%}).

A similar trend appears for \emph{Commonsense}. Katz \bottomk achieves the best score (57.8\%, \pos{0.5\%}), while Katz \topk substantially underperforms the baseline (56.1\%, \dec{1.2\%}). Betweenness \bottomk is roughly neutral (\pos{0.1\%}), while Betweenness \topk again slightly decreases performance (\dec{0.6\%}).

These results suggest that peripheral regions of the web contain useful long-tail and diverse knowledge signals that are beneficial for factual recall and commonsense reasoning tasks.


\paragraph{Structured reasoning benefits from both \topk and \bottomk sampling.}
Unlike the knowledge-oriented categories, \emph{Reasoning} improves under all centrality-based sampling strategies. Katz \topk achieves the strongest result (20.4\%\pos{1.4\%}), followed closely by Katz \bottomk (\pos{1.2\%}). Betweenness \topk and \bottomk produce similar gains (\pos{0.9\%} and \pos{0.8\%} respectively).

This indicates that reasoning tasks benefit from structural selection in general. However, the relatively stronger gains from \topk sampling methods suggest that highly influential hosts may contain more structured or procedural content useful for these tasks.

\paragraph{Comprehension and language understanding exhibit asymmetric behavior.}
For \emph{Comprehension}, \bottomk and \topk behave very differently. Katz \bottomk improves over baseline (\pos{0.7\%}), while Katz \topk substantially hurts performance (\dec{2.4\%}), the largest degradation in the table. A similar but weaker pattern appears for \emph{Language Understanding}, with only Katz \bottomk improving.

These results suggest that aggressive concentration on structurally central hosts may reduce linguistic diversity or contextual variability, which are important for comprehension-oriented tasks.

\paragraph{Centrality metric matters.}
The two centrality measures also behave differently. 
Katz centrality generally produces larger, less stable positive and negative shifts than Betweenness centrality, such as the strongest gains on \emph{Reasoning} (\pos{1.4\%}) but also the largest degradation on \emph{Comprehension} (\dec{2.4\%}), suggesting that recursive influence captures a stronger structural signal than shortest-path bridging.

\begin{table}[t]
\centering
\caption{Average accuracy across different task categories for \topk and \bottomk sampling at 1B scale. \textbf{Betw.} denotes Betweenness centrality scores. \textbf{Katz} denotes Katz centrality scores.}
\label{tab:knowledge_vs_reasoning_1b}
\small
\begin{tabular}{llllll}
\toprule
\textbf{Method} & \textbf{Commonsense} & \textbf{Comprehension} & \textbf{Knowledge} & \textbf{Reasoning} & \textbf{Language} \\
\midrule
Random     & 57.3 & 37.9 & 34.2 & 19.0 & 39.9 \\
Betw.\ \topk  & 56.7 \dec{0.6} & 37.1 \dec{0.8} & 33.9 \dec{0.3} & 19.9 \pos{0.9} & 39.2 \dec{0.7} \\
Betw.\ \bottomk & 57.4 \pos{0.1} & 36.9 \dec{1.0} & 35.4 \pos{1.2} & 19.8 \pos{0.8} & 39.5 \dec{0.4} \\
Katz \topk   & 56.1 \dec{1.2} & 35.5 \dec{2.4} & 34.4 \pos{0.2} & 20.4 \pos{1.4} & 39.1 \dec{0.8} \\
Katz \bottomk  & 57.8 \pos{0.5} & 38.6 \pos{0.7} & 35.3 \pos{1.1} & 20.2 \pos{1.2} & 40.2 \pos{0.3} \\
\bottomrule
\end{tabular}
\end{table}

\subsection{Centrality Score Distribution}
\label{sec:centrality_distribution}

\begin{figure}[t]
    \centering
    \includegraphics[width=0.81\linewidth]{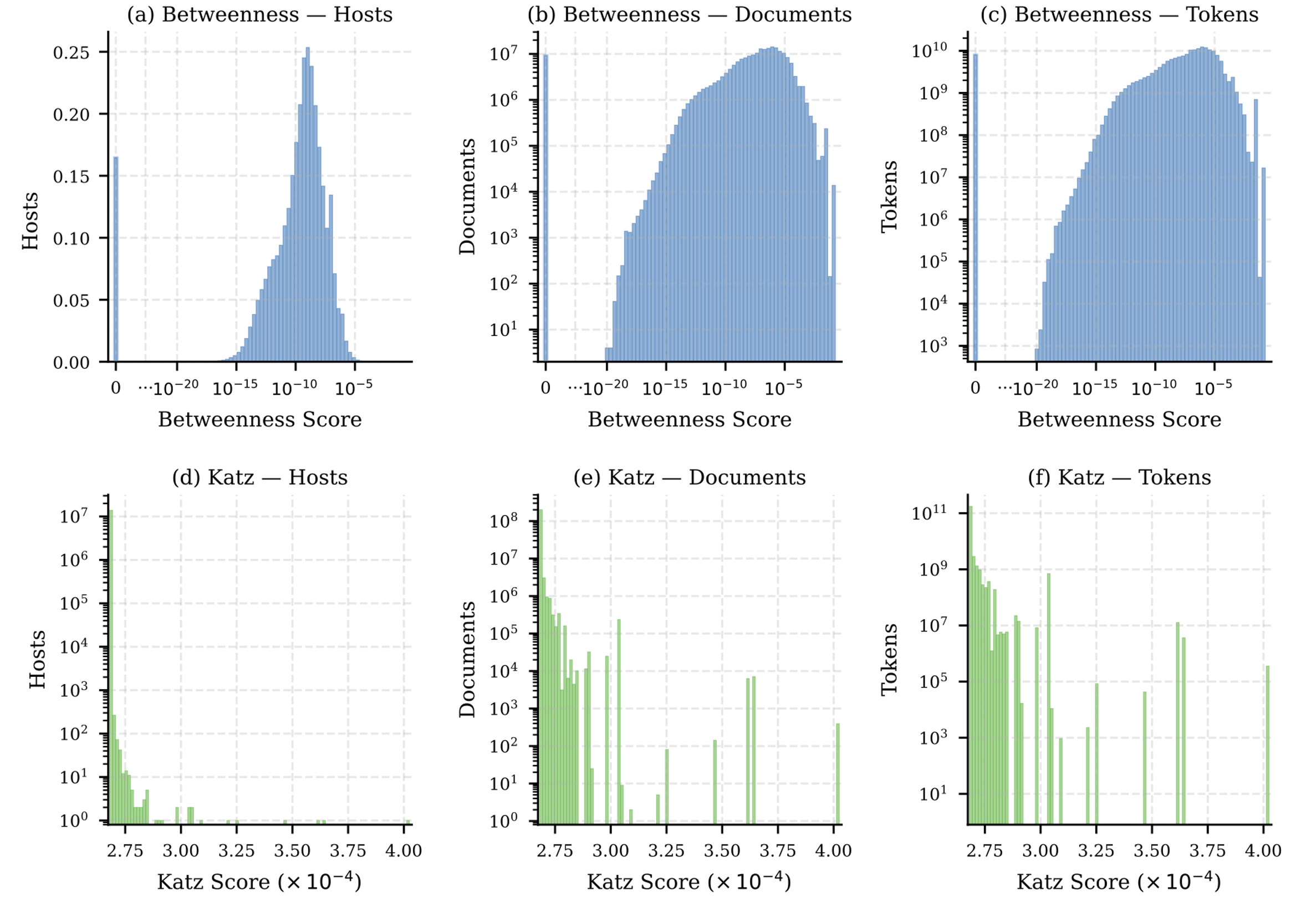}
    \caption{Histograms of Betweenness centrality scores and Katz centrality scores distribution, with respect to hosts, documents, and tokens. 
    }
    \label{fig:dist}
\end{figure}

Figure~\ref{fig:dist} shows the distributions of Betweenness and Katz centrality scores at three levels of aggregation: hosts only, weighted by documents per host, and weighted by tokens per host. The latter two reflect the effective score distribution over the training corpus, since each document inherits its \mbox{host's centrality score}.

The two metrics exhibit very different shapes. On a log scale, Betweenness (Fig.~\ref{fig:dist}a--c) is bell-shaped and roughly symmetric, with the bulk of mass between $10^{-15}$ and $10^{-5}$. A discrete spike at zero reflects a structural artifact of the Common Crawl host graph: it consists of roughly a dozen weakly connected components, and hosts in the smaller components receive near-zero betweenness because the shortest paths through them are bounded by their component size. Katz centrality (Fig.~\ref{fig:dist}d--f) is instead sharply right-skewed: most hosts cluster near the low end of the score range ($\sim 2.75 \times 10^{-4}$), with a long, sparse tail of high-scoring hosts that are recursively linked to other influential hosts. Document- and token-weighting shifts mass slightly toward higher scores in both cases, since central hosts contribute more content to the corpus, but the qualitative shapes are preserved.

\subsection{Mixture Sampling}

\begin{figure}
    \centering
    \vspace{-10pt}
    \includegraphics[width=0.9\linewidth]{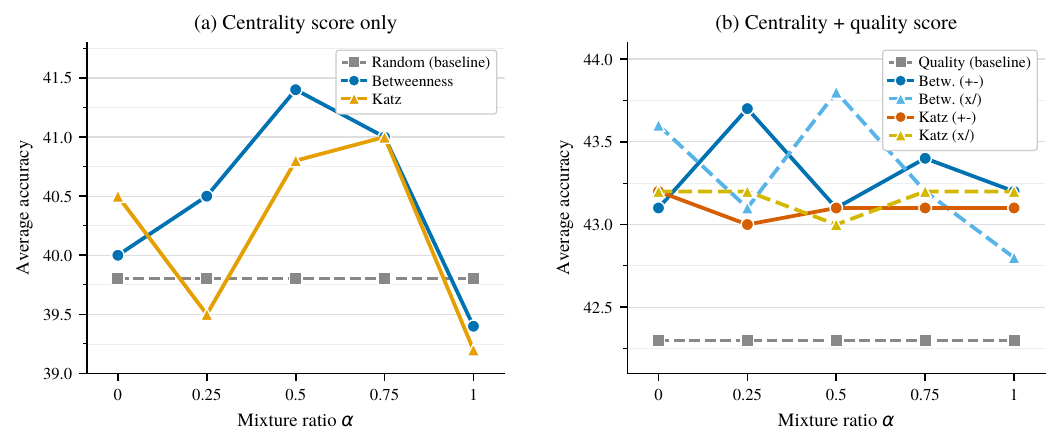}
    \vspace{-8pt}
    \caption{Average accuracy of mixture sampling with Betweenness centrality score and Katz centrality score at 1B scale, varying ratio of \topk and \bottomk tokens. (a) is centrality score only. (b) is centrality score combined with quality score. \opaddsub means combining with addition for \topk and subtraction for \bottomk. \opmuldiv means combining with multiplication for \topk and division~for~\bottomk.}
    \label{fig:centrality_mixture}
\end{figure}

We investigate the effect of mixing \topk and \bottomk sampling by varying the proportion of \topk and \bottomk documents in the sampled data. We find a mixture at around 1:1 yields the strongest performance.
\cref{fig:centrality_mixture}(a) summarizes the 23-task averages across mixture ratios \mbox{and centrality metrics}.

\paragraph{Mixing outperforms pure sampling.}
Neither pure \topk nor pure \bottomk achieves the gain of $\alpha=0.5$ mixture. This confirms our central hypothesis: central and peripheral web regions encode complementary capabilities, and balancing them yields better data mixture than either extreme alone.

\paragraph{The optimal ratio is roughly balanced.}
Across betweenness mixtures, $\alpha=0.5$ outperforms both $\alpha=0.25$ (40.5\%) and $\alpha=0.75$ (41.0\%). This suggests that neither central nor peripheral documents should dominate---the best pretraining data draws roughly equally from both structural extremes \mbox{of the web graph}.

Betweenness mixtures are consistently stronger than or equal to Katz mixtures at the 1B scale, with the gap largest at $\alpha=0.5$ (\pos{0.6\%}). There is a clear non-monotonic pattern for betweenness: performance peaks at $\alpha=0.5$ and declines on either side. This inverted-U shape supports the complementarity hypothesis---too much of either extreme hurts. Katz mixtures also support this trend, with performance improving as the proportion of central documents increases (39.5\% to 40.8\% to 41.0\%), but then decreasing to 39.2\% when there are too many (Katz \topk). This may reflect the different nature of Katz centrality, which emphasizes local connectivity rather than global bridging.

At 400M, mixture improvements are smaller but present. Katz $\alpha=0.25$ improves 0.1\% over the baseline while Betweenness $\alpha=0.75$ makes gains on specific tasks like ARC Easy (2.6\%), Winogrande (2.4\%), and ARC Challenge (2.5\%). This weaker signal is expected: smaller models have less capacity to leverage the complementary information from different web regions.

\subsection{Combining with Quality Scores}

\cref{fig:centrality_mixture}(b) reports results for combining centrality with quality scores at 1B scale. The headline finding is that centrality extracts robustly positive value \emph{on top of} the quality filter: every one of the 18 reported configurations exceeds the quality-only baseline of 42.3\%, with gains ranging from 0.5\% to 1.5\%. The strongest configuration, \emph{Multiply Betweenness 50\% Top, achieves an average of 43.8\%}---a 1.5\% improvement over quality-only and a 4.0\% improvement over random sampling. This indicates that web graph centrality is not merely competitive with content-based quality scoring
but consistently complementary to it. The signal centrality captures (structural position in the hyperlink graph) appears largely orthogonal to what DCLM-fasttext captures, so combining the two yields broadly compounding gains.

\section{Conclusion}

We introduced \ours, a lightweight pretraining data selection framework that uses structural position in the Common Crawl web graph as a signal for sampling documents. Our results show that different regions of the web graph encode complementary capabilities: structurally central hosts improve symbolic and procedural reasoning more, while peripheral hosts improve commonsense and factual knowledge more. Mixing these regions outperforms either extreme alone, and combining centrality with quality-based filtering yields further gains.
Unlike prior data selection methods that require auxiliary model training, influence estimation, or domain taxonomy construction, \ours computes centrality scores once over publicly available web graph using standard graph algorithms, requiring less than 9 GPU-hours. The resulting signal is lightweight, reusable, and complementary to existing content-based approaches, suggesting that web graph topology is a promising new axis for pretraining data curation.



\section*{Acknowledgments}
This research is partially funded by the National Science Foundation (IIS-2211779) and a Sloan Research Fellowship. This research is also supported by Princeton Language and Intelligence (PLI) and Princeton AI Lab. The experiments in this work were conducted on the Della high-performance computing cluster, as a part of Princeton Research Computing resources. 


\bibliography{ref}
\bibliographystyle{colm2026_conference}

\newpage
\appendix

\section{Experiment Details}
\label{app:exp}
After selecting documents according to our graph-based sampling strategy, we construct an untokenized dataset in JSONL format consistent with DCLM specifications. Tokenization and shuffling are performed using DCLM’s official Rust-based tokshuf pipeline. Specifically, we tokenize with the GPT‑NeoX tokenizer at sequence length 2049, following DCLM’s standard configuration. The Rust pipeline produces WebDataset shards and generates the corresponding manifest file required by the DCLM training script. For each experiment, we create a dataset reference JSON to integrate seamlessly with the DCLM workflow. We do not modify tokenizer settings, sequence length, sharding configuration, or preprocessing logic. By using the official tokenize-and-shuffle implementation, we maintain identical preprocessing behavior to prior DCLM submissions and eliminate potential implementation-induced variation.

Model training is executed using DCLM’s training.train entrypoint, which builds upon the OpenLM framework. All experiments follow fixed DCLM scale-specific recipes. We evaluate two compute scales: 400M‑1x, which trains a 412M-parameter model on approximately 8.2B tokens, and 1B‑1x, which trains a 1.4B-parameter model on approximately 28B tokens. For each scale, DCLM specifies the model architecture, number of layers, hidden size, attention heads, learning rate schedule, warmup steps, batch size, weight decay, gradient accumulation, and total number of training tokens. We use these configurations exactly as provided, without modification. In practice, we train with slightly more raw tokens than the nominal DCLM target to account for token loss during shuffling and padding, ensuring the effective training token count matches the intended compute budget. The 400M model takes around 20 hours on 4 H100 GPUs while the 1B model takes around 90 hours on 4~H100~GPUs.

\begin{table}[H]
\centering
\caption{DCLM CORE v2 evaluation tasks used in our experiments, along with their categories.}
\label{tab:eval-tasks}
\resizebox{\textwidth}{!}{%
\begin{tabular}{llcl}
\toprule
\textbf{Task} & \textbf{Category} & \textbf{Few-shot} & \textbf{Description} \\
\midrule
HellaSwag & Commonsense \& Reasoning & 0/10 & Sentence completion, grounded commonsense \\
CommonsenseQA & Commonsense \& Reasoning & 10 & 5-choice commonsense QA \\
COPA & Commonsense \& Reasoning & 0 & Causal reasoning, cause/effect \\
PIQA & Commonsense \& Reasoning & 10 & Physical commonsense (2-choice) \\
Winograd & Commonsense \& Reasoning & 0 & Pronoun resolution, commonsense \\
Winogrande & Commonsense \& Reasoning & 0 & Large-scale Winograd-style \\
\midrule
BoolQ & QA \& Comprehension & 10 & Binary yes/no QA from passages \\
SQuAD (v2) & QA \& Comprehension & 10 & Extractive QA; may be unanswerable \\
CoQA & QA \& Comprehension & 0 & Conversational QA \\
OpenBookQA & QA \& Comprehension & 0 & Multi-step reasoning + commonsense \\
\midrule
ARC Easy & Science \& Factual Knowledge & 10 & Grade-school science (easy), 4-choice \\
ARC Challenge & Science \& Factual Knowledge & 10 & Grade-school science (hard), 4-choice \\
Jeopardy & Science \& Factual Knowledge & 10 & Diverse trivia, generative \\
QA Wikidata & Science \& Factual Knowledge & 10 & Big-Bench factual completions \\
MMLU & Science \& Factual Knowledge & 5 & 57-subject academic QA (aggregate) \\
LSAT-AR & Science \& Factual Knowledge & 3 & Analytical reasoning from LSAT \\
\midrule
CS Algorithms & Symbolic \& Algo Reasoning & 10 & Big-Bench: recursion, DP execution \\
Dyck Languages & Symbolic \& Algo Reasoning & 10 & Big-Bench: balanced bracket completion \\
Operators & Symbolic \& Algo Reasoning & 10 & Big-Bench: novel operator definitions \\
Repeat Copy Logic & Symbolic \& Algo Reasoning & 10 & Big-Bench: words repeating and ordering \\
\midrule
LAMBADA & Language Understanding & 0 & Last-word prediction, long context \\
Language Identification & Language Understanding & 10 & Big-Bench: identify written language \\
\bottomrule
\end{tabular}
}
\end{table}

\begin{table}[H]
\centering
\caption{Licenses for existing assets used in this paper.}
\label{tab:licenses}
\small
\resizebox{\textwidth}{!}{%
\begin{tabular}{llll}
\toprule
\textbf{Asset} & \textbf{Type} & \textbf{License} & \textbf{URL} \\
\midrule
Common Crawl Web Graph & Data & Custom ToU\textsuperscript{\dag} & \url{https://commoncrawl.org/web-graphs} \\
cc-webgraph tools & Code & Apache 2.0 & \url{https://github.com/commoncrawl/cc-webgraph} \\
DCLM framework & Code & MIT & \url{https://github.com/mlfoundations/dclm} \\
DCLM data pool & Data & MIT\textsuperscript{\ddag} & \url{https://github.com/mlfoundations/dclm} \\
WebOrganizer Corpus-200B & Data & Not specified\textsuperscript{\S} & \url{https://huggingface.co/datasets/WebOrganizer/Corpus-200B} \\
RefinedWeb filters & Code & Apache 2.0 & \url{https://huggingface.co/datasets/tiiuae/falcon-refinedweb} \\
BFF deduplication & Code & Apache 2.0\textsuperscript{\P} & \url{https://github.com/allenai/bff} \\
cuGraph (RAPIDS) & Code & Apache 2.0 & \url{https://github.com/rapidsai/cugraph} \\
\bottomrule
\end{tabular}
}
\end{table}

\section{Full Results}
\label{app:results}
Pure centrality sampling at 1B scale tells a nuanced story. Bottom-$K$ outperforms Top-$K$ on average: Katz Bottom-$K$ achieves 0.405 (+0.4pp over baseline), while Katz Top-$K$ achieves only 0.392 ($-$0.9pp). This pattern holds across both centrality metrics.

At the smaller 400M scale, the pattern is weaker. The baseline (0.325) ties with Katz Bottom-$K$ (0.325) as the best average. Pure centrality strategies show less consistent improvement, likely because the 400M model has less capacity to exploit the structural signal. However, the same task-level asymmetry exists: Katz Top-$K$ achieves 0.601 on BoolQ (+7.3pp over baseline) while the baseline wins on bigbench\_qa\_wikidata (0.444 vs.\ 0.387, +5.7pp).

\begin{table}[H]
\centering
\caption{Accuracy on 23 tasks from DCLM CORE v2 benchmark. All 1.4B models are trained on 28B tokens selected by baseline methods and our \ours method. Our methods use the betweenness centrality scores with a Top-$K$ and Bottom-$K$ mixture of 1:1. We use multiplication for combining with the quality scores, denoted as \textbf{Ours+}. Note that while our \ours is close to WebOrganizer baseline, our method is significantly cheaper and more transferable.}
\label{tab:main}
\resizebox{\textwidth}{!}{
\scriptsize
\begin{tabular}{lccccccc}
\toprule
\textbf{Task} & \textbf{Random} & \textbf{Quality} & \textbf{WebOrg} & \textbf{WebOrg+} & \textbf{PageRank} & \textbf{Ours} & \textbf{Ours+} \\
\midrule
MMLU & 24.6 & 23.7 & 25.0 & 24.4 & 25.1 & \textbf{25.4} & 24.0 \\
HellaSwag (zero-shot) & 55.1 & 57.3 & 59.2 & \textbf{61.2} & 51.9 & 55.4 & 57.0 \\
Jeopardy & 15.3 & 26.8 & 19.4 & 24.6 & 15.0 & 13.0 & \textbf{27.1} \\
QA Wikidata & 58.9 & \textbf{60.2} & 58.9 & 59.0 & 58.4 & \textbf{60.2} & 59.8 \\
ARC Easy & 56.9 & 66.2 & 64.9 & \textbf{67.0} & 57.3 & 57.6 & 66.2 \\
ARC Challenge & 28.3 & \textbf{36.7} & 34.5 & \textbf{36.7} & 28.8 & 29.4 & 35.9 \\
COPA & 68.0 & \textbf{73.0} & 69.0 & 69.0 & 67.0 & 70.0 & 70.0 \\
CommonsenseQA & 24.1 & 21.9 & 22.5 & 32.3 & 29.2 & 31.8 & \textbf{32.5} \\
PIQA & 72.6 & 73.1 & 75.0 & \textbf{75.5} & 71.7 & 73.4 & 74.2 \\
OpenBookQA & 35.0 & 38.2 & 35.8 & 39.4 & 35.8 & 37.8 & \textbf{39.6} \\
LAMBADA & 55.3 & \textbf{60.3} & 51.4 & 52.2 & 51.5 & 55.0 & 58.9 \\
HellaSwag & 55.9 & 58.0 & 59.7 & \textbf{61.9} & 52.4 & 56.4 & 57.5 \\
Winograd & 71.4 & 76.9 & 73.6 & \textbf{77.3} & 70.7 & 72.2 & 76.6 \\
Winogrande & 54.1 & \textbf{58.4} & 57.9 & 56.3 & 55.6 & 57.4 & 58.1 \\
Dyck Languages & 15.8 & 21.2 & 19.1 & 22.5 & 16.8 & 23.1 & \textbf{25.4} \\
LSAT-AR & 21.3 & 20.0 & 25.2 & 22.6 & 23.9 & \textbf{27.0} & 25.2 \\
CS Algorithms & 42.3 & 42.7 & \textbf{44.9} & 42.9 & 41.3 & 41.2 & \textbf{44.9} \\
Operators & 17.1 & 19.0 & 19.5 & 19.0 & 19.0 & 18.1 & \textbf{20.0} \\
Repeat Copy Logic & 3.1 & 0 & \textbf{6.3} & 3.1 & 0 & 3.1 & 0 \\
SQuAD & 32.5 & 33.7 & 36.4 & 36.4 & 29.3 & 33.3 & \textbf{37.4} \\
CoQA & 26.0 & 29.5 & 27.8 & 29.7 & 24.2 & 26.3 & \textbf{30.8} \\
BoolQ & 58.0 & 50.8 & 56.9 & 59.9 & 60.2 & 60.2 & \textbf{62.4} \\
Language Identification & 24.5 & 25.2 & 25.2 & 25.3 & 24.7 & \textbf{25.4} & 24.8 \\
\midrule
\textbf{Average} & 39.8 & 42.3 & 42.1 & 43.4 & 39.6 & 41.4 & \textbf{43.8} \\
\bottomrule
\end{tabular}
}
\end{table}


\begin{table}[H]
\centering
\caption{Pure centrality sampling at 1B scale (1.4B parameters, 28B tokens). Each column selects documents whose hosts fall in the highest (Top-$K$) or lowest (Bottom-$K$) centrality stratum. Katz Bottom-$K$ achieves the highest average, suggesting peripheral web regions encode complementary capabilities at this scale.}
\label{tab:1b_pure}
\resizebox{\textwidth}{!}{
\scriptsize
\begin{tabular}{lcccccc}
\toprule
\textbf{Task} & \textbf{Baseline} & \textbf{Betw.\ Top-$K$} & \textbf{Betw.\ Bottom-$K$} & \textbf{Katz Top-$K$} & \textbf{Katz Bottom-$K$} \\
\midrule
mmlu\_fewshot & 0.246 & 0.245 & \textbf{0.257} & 0.248 & 0.250 \\
hellaswag\_zeroshot & \textbf{0.551} & 0.535 & 0.549 & 0.524 & 0.548 \\
jeopardy & 0.153 & 0.101 & 0.145 & 0.120 & \textbf{0.165} \\
bigbench\_qa\_wikidata & 0.589 & 0.596 & 0.595 & 0.608 & \textbf{0.612} \\
arc\_easy & 0.569 & 0.574 & \textbf{0.584} & 0.573 & 0.583 \\
arc\_challenge & 0.283 & 0.288 & 0.298 & \textbf{0.305} & 0.301 \\
copa & \textbf{0.680} & 0.660 & 0.670 & 0.660 & 0.650 \\
commonsense\_qa & 0.241 & 0.242 & 0.243 & 0.207 & \textbf{0.284} \\
piqa & 0.726 & 0.728 & \textbf{0.737} & 0.719 & 0.726 \\
openbook\_qa & 0.350 & 0.338 & 0.358 & 0.346 & \textbf{0.370} \\
lambada\_openai & \textbf{0.553} & 0.538 & 0.544 & 0.534 & 0.549 \\
hellaswag & \textbf{0.559} & 0.539 & 0.557 & 0.531 & 0.552 \\
winograd & \textbf{0.714} & 0.696 & 0.714 & 0.733 & 0.725 \\
winogrande & 0.541 & \textbf{0.578} & 0.551 & 0.554 & 0.560 \\
bigbench\_dyck\_languages & 0.158 & \textbf{0.213} & 0.195 & 0.165 & 0.181 \\
agi\_eval\_lsat\_ar & 0.213 & 0.226 & \textbf{0.248} & 0.213 & 0.204 \\
bigbench\_cs\_algorithms & 0.423 & 0.380 & 0.426 & \textbf{0.456} & 0.414 \\
bigbench\_operators & 0.171 & 0.171 & 0.171 & \textbf{0.195} & 0.181 \\
bigbench\_repeat\_copy\_logic & \textbf{0.031} & \textbf{0.031} & 0 & 0 & \textbf{0.031} \\
squad & \textbf{0.325} & 0.294 & 0.316 & 0.286 & 0.321 \\
coqa & \textbf{0.260} & 0.249 & 0.256 & 0.240 & 0.236 \\
boolq & 0.580 & 0.602 & 0.547 & 0.548 & \textbf{0.616} \\
bigbench\_language\_id & 0.245 & 0.247 & 0.247 & 0.247 & \textbf{0.254} \\
\midrule
\textbf{Average} & 0.398 & 0.394 & 0.400 & 0.392 & \textbf{0.405} \\
\bottomrule
\end{tabular}
}
\end{table}


\begin{table}[H]
\centering
\caption{Mixture sampling at 1B scale (1.4B parameters, 28B tokens). Each column combines a specified percentage of Top-$K$ (central) documents with the remainder drawn from Bottom-$K$ (peripheral) documents. Betweenness 50\% Top achieves the highest average (0.414), outperforming both the uniform baseline (0.398) and all pure sampling strategies from Table~\ref{tab:1b_pure}.}
\label{tab:1b_mixture}
\resizebox{\textwidth}{!}{
\scriptsize
\begin{tabular}{lccccccc}
\toprule
\textbf{Task} & \textbf{Baseline} & \textbf{Betw.\ 25\%} & \textbf{Betw.\ 50\%} & \textbf{Betw.\ 75\%} & \textbf{Katz 25\%} & \textbf{Katz 50\%} & \textbf{Katz 75\%} \\
\midrule
mmlu\_fewshot & 0.246 & 0.251 & \textbf{0.254} & 0.252 & 0.233 & 0.252 & 0.241 \\
hellaswag\_zeroshot & 0.551 & \textbf{0.567} & 0.554 & 0.543 & 0.542 & 0.557 & 0.547 \\
jeopardy & 0.153 & 0.153 & 0.130 & 0.137 & \textbf{0.175} & 0.156 & 0.149 \\
bigbench\_qa\_wikidata & 0.589 & 0.599 & 0.602 & \textbf{0.614} & 0.604 & 0.608 & 0.613 \\
arc\_easy & 0.569 & 0.585 & 0.576 & 0.574 & \textbf{0.600} & 0.585 & 0.578 \\
arc\_challenge & 0.283 & 0.295 & 0.294 & 0.294 & 0.303 & 0.288 & \textbf{0.304} \\
copa & 0.680 & 0.670 & 0.700 & 0.660 & 0.670 & \textbf{0.720} & 0.700 \\
commonsense\_qa & 0.241 & 0.201 & 0.318 & \textbf{0.342} & 0.201 & 0.229 & 0.279 \\
piqa & 0.726 & 0.734 & 0.734 & 0.733 & 0.731 & \textbf{0.738} & 0.736 \\
openbook\_qa & 0.350 & 0.352 & \textbf{0.378} & 0.358 & 0.340 & 0.336 & 0.360 \\
lambada\_openai & 0.553 & \textbf{0.555} & 0.550 & 0.530 & 0.552 & 0.554 & 0.553 \\
hellaswag & 0.559 & \textbf{0.577} & 0.564 & 0.551 & 0.557 & 0.561 & 0.554 \\
winograd & 0.714 & \textbf{0.769} & 0.722 & 0.751 & 0.751 & 0.736 & 0.733 \\
winogrande & 0.541 & 0.561 & \textbf{0.574} & \textbf{0.574} & 0.557 & 0.566 & 0.571 \\
bigbench\_dyck\_languages & 0.158 & 0.159 & 0.231 & \textbf{0.272} & 0.183 & 0.158 & 0.174 \\
agi\_eval\_lsat\_ar & 0.213 & 0.243 & 0.270 & 0.204 & 0.196 & 0.230 & \textbf{0.287} \\
bigbench\_cs\_algorithms & 0.423 & 0.448 & 0.412 & 0.417 & \textbf{0.457} & 0.442 & 0.446 \\
bigbench\_operators & 0.171 & 0.157 & 0.181 & \textbf{0.190} & 0.167 & 0.176 & 0.162 \\
bigbench\_repeat\_copy\_logic & 0.031 & 0.031 & 0.031 & 0 & 0.031 & \textbf{0.063} & 0.031 \\
squad & 0.325 & 0.328 & 0.333 & 0.322 & 0.307 & 0.329 & \textbf{0.351} \\
coqa & 0.260 & 0.268 & 0.263 & 0.255 & 0.259 & 0.244 & \textbf{0.272} \\
boolq & 0.580 & 0.563 & 0.602 & \textbf{0.610} & 0.430 & 0.593 & 0.554 \\
bigbench\_language\_id & 0.245 & 0.250 & 0.254 & \textbf{0.255} & 0.251 & 0.253 & 0.243 \\
\midrule
\textbf{Average} & 0.398 & 0.405 & \textbf{0.414} & 0.410 & 0.395 & 0.408 & 0.410 \\
\bottomrule
\end{tabular}
}
\end{table}


\begin{table}[H]
\centering
\caption{Pure centrality sampling at 400M scale (412M parameters, 8.2B tokens). The baseline (uniform sampling) achieves the highest average (0.325), with pure centrality strategies performing comparably. At this smaller scale, the signal from centrality alone does not consistently \mbox{outperform uniform sampling}.}
\label{tab:400m_pure}
\resizebox{\textwidth}{!}{
\scriptsize
\begin{tabular}{lcccccc}
\toprule
\textbf{Task} & \textbf{Baseline} & \textbf{Betw.\ Top-$K$} & \textbf{Betw.\ Bottom-$K$} & \textbf{Katz Top-$K$} & \textbf{Katz Bottom-$K$} \\
\midrule
mmlu\_fewshot & 0.247 & \textbf{0.255} & 0.242 & 0.248 & 0.255 \\
hellaswag\_zeroshot & 0.366 & 0.344 & \textbf{0.381} & 0.342 & 0.377 \\
jeopardy & 0.016 & \textbf{0.020} & 0.014 & 0.008 & 0.008 \\
bigbench\_qa\_wikidata & \textbf{0.444} & 0.421 & 0.392 & 0.387 & 0.380 \\
arc\_easy & 0.449 & 0.452 & \textbf{0.455} & 0.448 & 0.443 \\
arc\_challenge & 0.232 & 0.222 & 0.240 & \textbf{0.244} & 0.235 \\
copa & 0.620 & 0.580 & 0.570 & 0.620 & \textbf{0.670} \\
commonsense\_qa & 0.268 & 0.267 & 0.224 & 0.252 & \textbf{0.367} \\
piqa & 0.670 & 0.670 & \textbf{0.680} & 0.659 & 0.675 \\
openbook\_qa & 0.312 & 0.314 & 0.302 & 0.314 & \textbf{0.330} \\
lambada\_openai & 0.384 & 0.354 & \textbf{0.378} & 0.357 & 0.386 \\
hellaswag & 0.368 & 0.344 & \textbf{0.379} & 0.344 & 0.381 \\
winograd & 0.612 & 0.586 & 0.608 & 0.593 & \textbf{0.626} \\
winogrande & 0.519 & \textbf{0.521} & 0.517 & 0.507 & 0.502 \\
bigbench\_dyck\_languages & 0.123 & 0.137 & 0.126 & \textbf{0.147} & 0.092 \\
agi\_eval\_lsat\_ar & 0.222 & \textbf{0.252} & 0.239 & 0.222 & \textbf{0.252} \\
bigbench\_cs\_algorithms & 0.396 & \textbf{0.430} & 0.355 & 0.352 & 0.355 \\
bigbench\_operators & \textbf{0.148} & 0.105 & 0.110 & 0.110 & 0.110 \\
bigbench\_repeat\_copy\_logic & \textbf{0.063} & 0.031 & 0.031 & 0 & 0.031 \\
squad & \textbf{0.107} & 0.057 & 0.091 & 0.047 & 0.082 \\
coqa & 0.124 & 0.120 & \textbf{0.133} & 0.117 & 0.131 \\
boolq & 0.528 & 0.439 & \textbf{0.563} & \textbf{0.601} & 0.529 \\
bigbench\_language\_id & 0.253 & \textbf{0.254} & 0.250 & 0.244 & 0.246 \\
\midrule
\textbf{Average} & \textbf{0.325} & 0.311 & 0.316 & 0.312 & 0.325 \\
\bottomrule
\end{tabular}
}
\end{table}


\begin{table}[H]
\centering
\caption{Mixture sampling at 400M scale (412M parameters, 8.2B tokens). Katz 25\% Top achieves the highest average (0.326), marginally outperforming the uniform baseline (0.325). Betweenness 75\% Top also shows gains on several individual tasks, indicating that the complementary signal from mixing central and peripheral documents is present even at smaller model scales.}
\label{tab:400m_mixture}
\resizebox{\textwidth}{!}{
\scriptsize
\begin{tabular}{lccccccc}
\toprule
\textbf{Task} & \textbf{Baseline} & \textbf{Betw.\ 25\%} & \textbf{Betw.\ 50\%} & \textbf{Betw.\ 75\%} & \textbf{Katz 25\%} & \textbf{Katz 50\%} & \textbf{Katz 75\%} \\
\midrule
mmlu\_fewshot & 0.247 & 0.238 & 0.244 & 0.242 & 0.241 & \textbf{0.259} & 0.235 \\
hellaswag\_zeroshot & 0.366 & 0.368 & 0.359 & 0.358 & \textbf{0.370} & 0.361 & 0.352 \\
jeopardy & 0.016 & \textbf{0.021} & 0.012 & 0.021 & 0.016 & 0.016 & 0.018 \\
bigbench\_qa\_wikidata & \textbf{0.444} & 0.433 & \textbf{0.447} & 0.421 & 0.415 & 0.428 & 0.421 \\
arc\_easy & 0.449 & 0.461 & 0.463 & \textbf{0.475} & 0.450 & 0.458 & 0.453 \\
arc\_challenge & 0.232 & 0.230 & 0.235 & \textbf{0.257} & 0.255 & 0.233 & 0.244 \\
copa & \textbf{0.620} & 0.660 & 0.630 & 0.650 & \textbf{0.680} & 0.580 & 0.610 \\
commonsense\_qa & 0.268 & 0.239 & 0.215 & \textbf{0.292} & 0.259 & 0.272 & 0.270 \\
piqa & 0.670 & 0.667 & \textbf{0.678} & 0.665 & 0.669 & 0.660 & 0.666 \\
openbook\_qa & 0.312 & 0.306 & 0.296 & 0.302 & 0.308 & 0.310 & \textbf{0.320} \\
lambada\_openai & 0.384 & 0.387 & 0.376 & 0.383 & \textbf{0.396} & 0.380 & 0.380 \\
hellaswag & 0.368 & \textbf{0.369} & 0.362 & 0.356 & 0.368 & 0.359 & 0.355 \\
winograd & 0.612 & 0.582 & 0.608 & 0.582 & \textbf{0.630} & 0.623 & 0.601 \\
winogrande & 0.519 & 0.494 & 0.502 & \textbf{0.543} & 0.500 & 0.501 & 0.515 \\
bigbench\_dyck\_languages & 0.123 & 0.136 & 0.152 & \textbf{0.165} & 0.149 & 0.108 & 0.161 \\
agi\_eval\_lsat\_ar & 0.222 & 0.235 & 0.170 & 0.213 & \textbf{0.265} & 0.222 & 0.226 \\
bigbench\_cs\_algorithms & 0.396 & 0.429 & 0.361 & 0.393 & 0.389 & \textbf{0.436} & 0.394 \\
bigbench\_operators & \textbf{0.148} & 0.105 & 0.133 & 0.124 & 0.143 & 0.124 & 0.110 \\
bigbench\_repeat\_copy\_logic & \textbf{0.063} & 0.031 & 0 & 0.031 & 0 & 0 & 0 \\
squad & 0.107 & \textbf{0.112} & 0.071 & 0.062 & 0.089 & 0.081 & 0.083 \\
coqa & 0.124 & 0.133 & 0.127 & \textbf{0.133} & \textbf{0.137} & 0.123 & 0.120 \\
boolq & 0.528 & 0.528 & 0.559 & \textbf{0.554} & 0.524 & \textbf{0.594} & 0.546 \\
bigbench\_language\_id & \textbf{0.253} & \textbf{0.253} & 0.247 & 0.245 & 0.247 & 0.249 & 0.247 \\
\midrule
\textbf{Average} & 0.325 & 0.323 & 0.315 & 0.325 & \textbf{0.326} & 0.321 & 0.319 \\
\bottomrule
\end{tabular}
}
\end{table}


\begin{table}[H]
\centering
\caption{Additive quality--centrality combination at 400M scale (412M parameters, 8.2B tokens). Normalized quality and centrality scores are summed, and documents are ranked by the combined score. Add Katz Top-$K$ achieves the highest average (0.351), substantially outperforming both the uniform baseline (0.325) and the quality-only filter (0.345), demonstrating that structural centrality provides an additive signal on top of content-based quality scoring.}
\label{tab:400m_additive}
\resizebox{\textwidth}{!}{
\scriptsize
\begin{tabular}{lcccccc}
\toprule
\textbf{Task} & \textbf{Baseline} & \textbf{Quality} & \textbf{Add Betw.\ Top} & \textbf{Add Betw.\ Bot.} & \textbf{Add Katz Top} & \textbf{Add Katz Bot.} \\
\midrule
mmlu\_fewshot & 0.247 & \textbf{0.253} & 0.244 & 0.252 & 0.250 & 0.245 \\
hellaswag\_zeroshot & 0.366 & 0.372 & 0.369 & 0.344 & \textbf{0.373} & 0.342 \\
jeopardy & 0.016 & \textbf{0.060} & 0.050 & 0.007 & 0.050 & 0.004 \\
bigbench\_qa\_wikidata & \textbf{0.444} & 0.400 & 0.396 & 0.391 & 0.399 & 0.394 \\
arc\_easy & 0.449 & \textbf{0.562} & 0.546 & 0.398 & 0.546 & 0.389 \\
arc\_challenge & 0.232 & 0.285 & 0.285 & 0.225 & \textbf{0.289} & 0.216 \\
copa & \textbf{0.620} & 0.600 & 0.600 & 0.560 & 0.600 & 0.610 \\
commonsense\_qa & 0.268 & 0.360 & 0.202 & 0.280 & \textbf{0.368} & 0.251 \\
piqa & 0.670 & 0.663 & \textbf{0.672} & 0.653 & 0.664 & 0.668 \\
openbook\_qa & 0.312 & 0.318 & 0.324 & 0.292 & \textbf{0.332} & 0.272 \\
lambada\_openai & 0.384 & \textbf{0.419} & 0.413 & 0.291 & 0.417 & 0.294 \\
hellaswag & 0.368 & 0.374 & 0.369 & 0.342 & \textbf{0.376} & 0.338 \\
winograd & 0.612 & 0.619 & 0.615 & 0.560 & \textbf{0.637} & 0.549 \\
winogrande & 0.519 & 0.515 & 0.507 & \textbf{0.523} & 0.507 & \textbf{0.540} \\
bigbench\_dyck\_languages & 0.123 & 0.259 & 0.226 & 0.132 & \textbf{0.278} & 0.104 \\
agi\_eval\_lsat\_ar & 0.222 & 0.248 & 0.183 & 0.191 & \textbf{0.274} & 0.191 \\
bigbench\_cs\_algorithms & 0.396 & 0.366 & 0.361 & 0.396 & \textbf{0.412} & 0.364 \\
bigbench\_operators & \textbf{0.148} & \textbf{0.171} & 0.167 & 0.133 & 0.157 & 0.148 \\
bigbench\_repeat\_copy\_logic & \textbf{0.063} & 0 & 0.031 & 0 & 0.031 & 0 \\
squad & 0.107 & 0.126 & 0.136 & 0.066 & \textbf{0.148} & 0.047 \\
coqa & 0.124 & 0.163 & 0.156 & 0.106 & \textbf{0.165} & 0.092 \\
boolq & 0.528 & 0.560 & 0.565 & \textbf{0.581} & 0.554 & 0.580 \\
bigbench\_language\_id & 0.253 & 0.247 & \textbf{0.261} & 0.250 & 0.254 & 0.251 \\
\midrule
\textbf{Average} & 0.325 & 0.345 & 0.334 & 0.303 & \textbf{0.351} & 0.300 \\
\bottomrule
\end{tabular}
}
\end{table}


\begin{table}[H]
\centering
\caption{Multiplicative quality--centrality combination at 400M scale (412M parameters, 8.2B tokens). Normalized quality and centrality scores are multiplied, and documents are ranked by the product. Both Multiply Betweenness Top-$K$ and Multiply Katz Top-$K$ achieve a tied best average of 0.345, matching the quality-only baseline. Bottom-$K$ variants underperform, confirming that multiplicative combination is most effective when selecting structurally central documents.}
\label{tab:400m_multiply}
\resizebox{\textwidth}{!}{
\scriptsize
\begin{tabular}{lcccccc}
\toprule
\textbf{Task} & \textbf{Baseline} & \textbf{Quality} & \textbf{Mult.\ Betw.\ Top} & \textbf{Mult.\ Betw.\ Bot.} & \textbf{Mult.\ Katz Top} & \textbf{Mult.\ Katz Bot.} \\
\midrule
mmlu\_fewshot & 0.247 & 0.253 & 0.242 & 0.251 & \textbf{0.258} & 0.248 \\
hellaswag\_zeroshot & 0.366 & \textbf{0.372} & 0.370 & 0.340 & 0.370 & 0.337 \\
jeopardy & 0.016 & \textbf{0.060} & 0.055 & 0.008 & 0.058 & 0.002 \\
bigbench\_qa\_wikidata & \textbf{0.444} & 0.400 & 0.420 & 0.359 & 0.402 & 0.371 \\
arc\_easy & 0.449 & \textbf{0.562} & 0.544 & 0.386 & \textbf{0.567} & 0.388 \\
arc\_challenge & 0.232 & \textbf{0.285} & 0.270 & 0.228 & 0.276 & 0.206 \\
copa & \textbf{0.620} & 0.600 & 0.590 & 0.560 & 0.600 & 0.590 \\
commonsense\_qa & 0.268 & \textbf{0.360} & 0.229 & 0.276 & 0.304 & 0.268 \\
piqa & 0.670 & 0.663 & 0.667 & 0.647 & \textbf{0.671} & 0.657 \\
openbook\_qa & 0.312 & 0.318 & 0.314 & 0.286 & \textbf{0.340} & 0.278 \\
lambada\_openai & 0.384 & \textbf{0.419} & \textbf{0.425} & 0.297 & 0.419 & 0.294 \\
hellaswag & 0.368 & \textbf{0.374} & 0.370 & 0.342 & 0.368 & 0.338 \\
winograd & 0.612 & 0.619 & 0.630 & 0.590 & \textbf{0.634} & 0.601 \\
winogrande & 0.519 & 0.515 & 0.519 & \textbf{0.530} & 0.520 & 0.523 \\
bigbench\_dyck\_languages & 0.123 & \textbf{0.259} & 0.251 & 0.124 & 0.238 & 0.136 \\
agi\_eval\_lsat\_ar & 0.222 & 0.248 & 0.265 & 0.204 & \textbf{0.283} & 0.178 \\
bigbench\_cs\_algorithms & 0.396 & 0.366 & \textbf{0.465} & 0.362 & 0.418 & 0.364 \\
bigbench\_operators & 0.148 & \textbf{0.171} & 0.162 & 0.162 & 0.148 & 0.152 \\
bigbench\_repeat\_copy\_logic & \textbf{0.063} & 0 & 0 & 0.031 & 0 & \textbf{0.063} \\
squad & 0.107 & 0.126 & 0.150 & 0.064 & \textbf{0.152} & 0.046 \\
coqa & 0.124 & \textbf{0.163} & 0.158 & 0.114 & 0.158 & 0.122 \\
boolq & 0.528 & 0.560 & 0.592 & 0.546 & 0.462 & \textbf{0.598} \\
bigbench\_language\_id & 0.253 & 0.247 & 0.251 & \textbf{0.258} & 0.248 & 0.251 \\
\midrule
\textbf{Average} & 0.325 & \textbf{0.345} & \textbf{0.345} & 0.303 & 0.343 & 0.305 \\
\bottomrule
\end{tabular}
}
\end{table}

\begin{table}[H]
\centering
\caption{Additive quality--betweenness centrality combination at 1B scale (1.4B parameters, 28B tokens). Normalized quality and betweenness centrality scores are summed, and documents are ranked by the combined score. Add Betw.\ 25\% achieves the highest average (0.437), outperforming both the uniform baseline (0.398) and the quality-only filter (0.423), demonstrating that betweenness centrality provides an additive signal on top of content-based quality scoring.}
\label{tab:1b_additive_betw}
\resizebox{\textwidth}{!}{
\scriptsize
\begin{tabular}{lcccccc}
\toprule
\textbf{Task} & \textbf{Baseline} & \textbf{Quality} & \textbf{Add Betw.\ 25\%} & \textbf{Add Betw.\ 50\%} & \textbf{Add Betw.\ 75\%} & \textbf{Add Betw.\ Bot.} \\
\midrule
mmlu\_fewshot & 0.246 & 0.237 & 0.251 & 0.251 & 0.257 & 0.258 \\
hellaswag\_zeroshot & 0.551 & 0.573 & 0.569 & 0.570 & 0.572 & 0.572 \\
jeopardy & 0.153 & 0.268 & 0.268 & \textbf{0.286} & 0.279 & 0.252 \\
bigbench\_qa\_wikidata & 0.589 & \textbf{0.602} & 0.588 & 0.600 & 0.592 & 0.595 \\
arc\_easy & 0.569 & 0.662 & 0.675 & 0.670 & 0.665 & 0.647 \\
arc\_challenge & 0.283 & 0.367 & \textbf{0.378} & 0.372 & 0.361 & 0.360 \\
copa & 0.680 & \textbf{0.730} & 0.660 & 0.710 & 0.710 & 0.710 \\
commonsense\_qa & 0.241 & 0.219 & 0.310 & 0.244 & 0.313 & 0.291 \\
piqa & 0.726 & 0.731 & \textbf{0.752} & 0.740 & 0.743 & 0.740 \\
openbook\_qa & 0.350 & 0.382 & 0.382 & 0.376 & 0.388 & 0.384 \\
lambada\_openai & 0.553 & 0.603 & 0.598 & 0.592 & 0.598 & \textbf{0.607} \\
hellaswag & 0.559 & 0.580 & 0.577 & 0.579 & 0.579 & 0.577 \\
winograd & 0.714 & 0.769 & \textbf{0.784} & 0.747 & 0.733 & 0.766 \\
winogrande & 0.541 & 0.584 & 0.597 & 0.573 & 0.593 & 0.577 \\
bigbench\_dyck\_languages & 0.158 & 0.212 & 0.196 & 0.159 & 0.167 & \textbf{0.201} \\
agi\_eval\_lsat\_ar & 0.213 & 0.200 & 0.243 & 0.243 & 0.239 & 0.217 \\
bigbench\_cs\_algorithms & 0.423 & 0.427 & 0.377 & \textbf{0.455} & 0.373 & 0.447 \\
bigbench\_operators & 0.171 & 0.190 & 0.200 & 0.210 & 0.210 & 0.214 \\
bigbench\_repeat\_copy\_logic & 0.031 & 0 & \textbf{0.094} & 0.031 & 0 & 0.063 \\
squad & 0.325 & 0.337 & 0.365 & 0.380 & \textbf{0.413} & 0.373 \\
coqa & 0.260 & 0.295 & 0.307 & 0.307 & 0.305 & 0.300 \\
boolq & 0.580 & 0.508 & 0.622 & 0.572 & \textbf{0.628} & 0.513 \\
bigbench\_language\_id & 0.245 & 0.252 & 0.249 & \textbf{0.254} & \textbf{0.255} & \textbf{0.254} \\
\midrule
\textbf{Average} & 0.398 & 0.423 & \textbf{0.437} & 0.431 & 0.434 & 0.431 \\
\bottomrule
\end{tabular}
}
\end{table}

\begin{table}[H]
\centering
\caption{Additive quality--Katz centrality combination at 1B scale (1.4B parameters, 28B tokens). Normalized quality and Katz centrality scores are summed, and documents are ranked by the combined score. All Katz variants achieve comparable average scores around 0.431--0.432, outperforming the uniform baseline (0.398) and approaching the quality-only filter (0.423), demonstrating that Katz centrality provides a consistent additive signal across sampling thresholds.}
\label{tab:1b_additive_katz}
\resizebox{\textwidth}{!}{
\scriptsize
\begin{tabular}{lcccccc}
\toprule
\textbf{Task} & \textbf{Baseline} & \textbf{Quality} & \textbf{Add Katz 25\%} & \textbf{Add Katz 75\%} & \textbf{Add Katz Bot.} & \textbf{Add Katz Top} \\
\midrule
mmlu\_fewshot & 0.246 & 0.237 & 0.248 & 0.242 & 0.244 & \textbf{0.266} \\
hellaswag\_zeroshot & 0.551 & 0.573 & 0.572 & 0.569 & \textbf{0.576} & 0.573 \\
jeopardy & 0.153 & 0.268 & 0.274 & 0.268 & 0.260 & 0.253 \\
bigbench\_qa\_wikidata & 0.589 & \textbf{0.602} & 0.590 & 0.587 & 0.577 & 0.598 \\
arc\_easy & 0.569 & 0.662 & 0.665 & \textbf{0.681} & 0.650 & 0.653 \\
arc\_challenge & 0.283 & 0.367 & 0.364 & 0.362 & 0.366 & 0.348 \\
copa & 0.680 & \textbf{0.730} & 0.700 & 0.670 & 0.700 & 0.670 \\
commonsense\_qa & 0.241 & 0.219 & 0.202 & 0.265 & \textbf{0.321} & 0.271 \\
piqa & 0.726 & 0.731 & 0.729 & 0.734 & 0.735 & \textbf{0.748} \\
openbook\_qa & 0.350 & 0.382 & 0.392 & 0.384 & \textbf{0.394} & 0.378 \\
lambada\_openai & 0.553 & 0.603 & 0.596 & 0.594 & 0.598 & 0.595 \\
hellaswag & 0.559 & 0.580 & 0.578 & 0.576 & \textbf{0.581} & 0.579 \\
winograd & 0.714 & 0.769 & 0.769 & 0.777 & 0.736 & 0.747 \\
winogrande & 0.541 & 0.584 & 0.586 & 0.581 & 0.598 & \textbf{0.602} \\
bigbench\_dyck\_languages & 0.158 & 0.212 & 0.161 & 0.188 & 0.179 & 0.194 \\
agi\_eval\_lsat\_ar & 0.213 & 0.200 & \textbf{0.287} & 0.183 & 0.200 & 0.213 \\
bigbench\_cs\_algorithms & 0.423 & 0.427 & 0.412 & 0.421 & 0.451 & 0.447 \\
bigbench\_operators & 0.171 & 0.190 & 0.214 & \textbf{0.238} & 0.186 & 0.210 \\
bigbench\_repeat\_copy\_logic & 0.031 & 0 & 0 & 0.031 & 0.063 & 0.031 \\
squad & 0.325 & 0.337 & 0.389 & 0.391 & 0.358 & 0.384 \\
coqa & 0.260 & 0.295 & 0.309 & 0.305 & 0.300 & \textbf{0.317} \\
boolq & 0.580 & 0.508 & 0.593 & 0.610 & 0.615 & 0.580 \\
bigbench\_language\_id & 0.245 & 0.252 & 0.250 & 0.250 & 0.247 & 0.249 \\
\midrule
\textbf{Average} & 0.398 & 0.423 & 0.430 & 0.431 & 0.432 & 0.431 \\
\bottomrule
\end{tabular}
}
\end{table}

\begin{table}[H]
\centering
\caption{Multiplicative quality--betweenness centrality combination at 1B scale (1.4B parameters, 28B tokens). Normalized quality and betweenness centrality scores are multiplied, and documents are ranked by the product. Mult.\ Betw.\ 50\% achieves the highest average (0.438), outperforming both the uniform baseline (0.398) and the quality-only filter (0.423). Bottom-$K$ variants remain competitive at this scale, unlike the 400M setting.}
\label{tab:1b_multiply_betw}
\resizebox{\textwidth}{!}{
\scriptsize
\begin{tabular}{lccccccc}
\toprule
\textbf{Task} & \textbf{Baseline} & \textbf{Quality} & \textbf{Mult.\ Betw.\ 25\%} & \textbf{Mult.\ Betw.\ 50\%} & \textbf{Mult.\ Betw.\ 75\%} & \textbf{Mult.\ Betw.\ Bot.} & \textbf{Mult.\ Betw.\ Top} \\
\midrule
mmlu\_fewshot & 0.246 & 0.237 & 0.248 & 0.240 & 0.236 & 0.236 & 0.250 \\
hellaswag\_zeroshot & 0.551 & 0.573 & 0.574 & 0.570 & \textbf{0.576} & 0.572 & 0.573 \\
jeopardy & 0.153 & 0.268 & 0.278 & 0.271 & 0.272 & 0.245 & 0.257 \\
bigbench\_qa\_wikidata & 0.589 & 0.602 & 0.586 & 0.598 & 0.599 & 0.597 & \textbf{0.606} \\
arc\_easy & 0.569 & 0.662 & 0.672 & 0.662 & 0.662 & 0.650 & 0.647 \\
arc\_challenge & 0.283 & 0.367 & \textbf{0.376} & 0.359 & 0.371 & 0.340 & 0.345 \\
copa & 0.680 & 0.730 & 0.680 & 0.700 & 0.680 & 0.680 & 0.710 \\
commonsense\_qa & 0.241 & 0.219 & 0.265 & 0.325 & 0.258 & \textbf{0.342} & 0.220 \\
piqa & 0.726 & 0.731 & 0.725 & \textbf{0.742} & 0.736 & 0.743 & 0.742 \\
openbook\_qa & 0.350 & 0.382 & 0.376 & 0.396 & \textbf{0.398} & 0.388 & 0.388 \\
lambada\_openai & 0.553 & 0.603 & 0.592 & 0.589 & 0.595 & 0.594 & 0.592 \\
hellaswag & 0.559 & 0.580 & 0.580 & 0.575 & 0.581 & 0.577 & \textbf{0.582} \\
winograd & 0.714 & 0.769 & \textbf{0.777} & 0.766 & 0.769 & 0.744 & 0.769 \\
winogrande & 0.541 & 0.584 & 0.574 & 0.581 & \textbf{0.594} & 0.590 & 0.581 \\
bigbench\_dyck\_languages & 0.158 & 0.212 & 0.204 & 0.254 & 0.180 & \textbf{0.285} & 0.235 \\
agi\_eval\_lsat\_ar & 0.213 & 0.200 & 0.239 & \textbf{0.252} & 0.239 & 0.226 & 0.243 \\
bigbench\_cs\_algorithms & 0.423 & 0.427 & 0.398 & \textbf{0.449} & 0.399 & 0.431 & 0.357 \\
bigbench\_operators & 0.171 & 0.190 & 0.186 & 0.200 & 0.200 & 0.205 & \textbf{0.224} \\
bigbench\_repeat\_copy\_logic & 0.031 & 0 & 0.031 & 0 & 0.031 & 0.031 & \textbf{0.063} \\
squad & 0.325 & 0.337 & 0.381 & 0.374 & \textbf{0.393} & 0.373 & 0.375 \\
coqa & 0.260 & 0.295 & 0.302 & 0.308 & \textbf{0.320} & 0.293 & 0.303 \\
boolq & 0.580 & 0.508 & 0.617 & 0.624 & 0.608 & \textbf{0.631} & 0.549 \\
bigbench\_language\_id & 0.245 & 0.252 & \textbf{0.255} & 0.248 & 0.248 & \textbf{0.255} & 0.245 \\
\midrule
\textbf{Average} & 0.398 & 0.423 & 0.431 & \textbf{0.438} & 0.432 & 0.436 & 0.428 \\
\bottomrule
\end{tabular}
}
\end{table}

\begin{table}[H]
\centering
\caption{Multiplicative quality--Katz centrality combination at 1B scale (1.4B parameters, 28B tokens). Normalized quality and Katz centrality scores are multiplied, and documents are ranked by the product. All Katz variants achieve comparable average scores around 0.430--0.432, outperforming the uniform baseline (0.398) and the quality-only filter (0.423). Bottom-$K$ variants remain competitive at this scale, unlike the 400M setting.}
\label{tab:1b_multiply_katz}
\resizebox{\textwidth}{!}{
\scriptsize
\begin{tabular}{lccccccc}
\toprule
\textbf{Task} & \textbf{Baseline} & \textbf{Quality} & \textbf{Mult.\ Katz 25\%} & \textbf{Mult.\ Katz 50\%} & \textbf{Mult.\ Katz 75\%} & \textbf{Mult.\ Katz Bot.} & \textbf{Mult.\ Katz Top} \\
\midrule
mmlu\_fewshot & 0.246 & 0.237 & 0.249 & 0.254 & \textbf{0.260} & 0.253 & 0.264 \\
hellaswag\_zeroshot & 0.551 & 0.573 & 0.572 & 0.570 & 0.572 & \textbf{0.574} & 0.572 \\
jeopardy & 0.153 & 0.268 & \textbf{0.286} & 0.273 & 0.270 & 0.263 & 0.261 \\
bigbench\_qa\_wikidata & 0.589 & 0.602 & 0.586 & 0.581 & 0.567 & \textbf{0.600} & 0.591 \\
arc\_easy & 0.569 & 0.662 & 0.669 & 0.668 & \textbf{0.666} & 0.648 & 0.656 \\
arc\_challenge & 0.283 & 0.367 & 0.369 & 0.353 & \textbf{0.378} & 0.346 & 0.356 \\
copa & 0.680 & 0.730 & 0.680 & 0.690 & \textbf{0.770} & 0.690 & 0.680 \\
commonsense\_qa & 0.241 & 0.219 & 0.261 & 0.363 & \textbf{0.382} & 0.216 & 0.277 \\
piqa & 0.726 & 0.731 & \textbf{0.744} & 0.736 & 0.740 & 0.737 & 0.733 \\
openbook\_qa & 0.350 & 0.382 & 0.388 & 0.396 & \textbf{0.408} & 0.384 & 0.382 \\
lambada\_openai & 0.553 & 0.603 & 0.594 & 0.599 & 0.594 & \textbf{0.600} & 0.589 \\
hellaswag & 0.559 & 0.580 & 0.581 & 0.576 & 0.580 & \textbf{0.583} & 0.580 \\
winograd & 0.714 & 0.769 & 0.755 & 0.762 & 0.740 & \textbf{0.791} & 0.758 \\
winogrande & 0.541 & 0.584 & \textbf{0.601} & 0.560 & 0.578 & 0.582 & 0.572 \\
bigbench\_dyck\_languages & 0.158 & 0.212 & 0.195 & 0.182 & 0.209 & \textbf{0.223} & 0.175 \\
agi\_eval\_lsat\_ar & 0.213 & 0.200 & 0.204 & 0.213 & 0.230 & \textbf{0.283} & 0.243 \\
bigbench\_cs\_algorithms & 0.423 & 0.427 & \textbf{0.455} & 0.355 & 0.372 & 0.386 & 0.439 \\
bigbench\_operators & 0.171 & 0.190 & \textbf{0.210} & 0.181 & 0.190 & 0.190 & 0.162 \\
bigbench\_repeat\_copy\_logic & 0.031 & 0 & 0 & 0.031 & 0.031 & 0.031 & \textbf{0.094} \\
squad & 0.325 & 0.337 & 0.391 & 0.383 & 0.365 & 0.383 & 0.376 \\
coqa & 0.260 & 0.295 & 0.313 & 0.310 & 0.316 & 0.304 & 0.300 \\
boolq & 0.580 & 0.508 & 0.593 & 0.606 & 0.464 & 0.607 & \textbf{0.625} \\
bigbench\_language\_id & 0.245 & 0.252 & 0.240 & 0.246 & 0.253 & 0.254 & \textbf{0.254} \\
\midrule
\textbf{Average} & 0.398 & 0.423 & 0.432 & 0.430 & 0.432 & 0.432 & 0.432 \\
\bottomrule
\end{tabular}
}
\end{table}

\paragraph{Best combination strategy shifts with scale.} As a secondary observation, we note that the best combination strategy reverses between the two scales: at 400M, Add Katz Top was strongest, while at 1B, Multiply Betweenness 50\% takes over. This may partially be attributed to the differing selectivity of the two strategies. Multiplicative scoring strongly suppresses documents that are low on either signal, while additive scoring is more permissive. At 400M, where the model has limited capacity, the broader signal from additive combination appears more useful; at 1B, the sharper selectivity of multiplicative combination yields better results. While interesting, this reversal is less practically important than the broader finding that \emph{both} combination strategies, in nearly all configurations, extract real value from centrality on top of quality filtering.

\section{Example Pretraining Documents}
\label{app:snippet}

\begin{table}[H]
\centering
\caption{%
  \textbf{Top hosts by betweenness centrality score.}
  The ten highest-scoring hosts from the web graph,
  with a representative URL snippet for each.
  Scores are computed over the host-level graph and reported in scientific notation.
}
\label{tab:betweenness_top}
\small
\setlength{\tabcolsep}{4pt}
\begin{tabular}{llp{7cm}}
\toprule
\textbf{Host} & \textbf{Score} & \textbf{Representative Snippet} \\
\midrule
\texttt{facebook.com}     & $1.98 \times 10^{-1}$ & Saving your new profile picture \ldots\ Wassup guys! At last we've prepared for you some cool stuff! Tacit Fury's brand new merch! \\[4pt]
\texttt{google.com}       & $1.18 \times 10^{-1}$ & Kellogg Company (Public, NYSE:K) Watch this stock Find more results for k +1.48 (2.18\%) Real-time: 3:21PM EST NYSE real-time data \ldots \\[4pt]
\texttt{youtube.com}      & $1.07 \times 10^{-1}$ & Oval Office Underdogs: The Poet Prophet. The interactive transcript could not be loaded. Rating is available when the video has been rented. \\[4pt]
\texttt{instagram.com}    & $5.94 \times 10^{-2}$ & Enter the Lane Bryant Makeover My Mom Contest thru May 4 for your chance to win a head-to-toe Lane Bryant Makeover \ldots \\[4pt]
\texttt{toptohigh.com}    & $4.10 \times 10^{-2}$ & Business to business marketing, commonly known as b2b marketing, refers to the interaction and marketing methods used to connect various businesses \ldots \\[4pt]
\texttt{linkedin.com}     & $3.64 \times 10^{-2}$ & Kristi Kaylor, Beverly Hills, California. 1. 6126 LLC, 2. 2 LOVE, 3. Mblem by Mandy Moore. Recommendations: 3 people have recommended Kristi. \\[4pt]
\texttt{articlement.com}  & $3.58 \times 10^{-2}$ & Silver Mountain Express is a private shuttle \& car service from Denver to Vail, Colorado that offers the perfect transportation solution \ldots \\[4pt]
\texttt{gmpg.org}         & $3.54 \times 10^{-2}$ & XFN: Getting Started. Join the XHTML Friends Network in four easy steps. 1. Pick one or more pages to make XFN Friendly \ldots \\[4pt]
\texttt{kingranks.com}    & $2.73 \times 10^{-2}$ & From crux gammata to swastika. What was possibly the most significant event of the 20th century, the Second World War, would not have occurred without the power of branding \ldots \\[4pt]
\texttt{en.wikipedia.org} & $2.46 \times 10^{-2}$ & Telluric contamination is contamination of the astronomical spectra by the Earth's atmosphere. Interference \ldots \\
\bottomrule
\end{tabular}
\end{table}


\begin{table}[H]
\centering
\caption{%
  \textbf{Bottom hosts by betweenness centrality score.}
  The ten lowest-scoring hosts from the web graph,
  with a representative URL snippet for each.
  Scores are computed over the host-level graph and reported in scientific notation.
}
\label{tab:betweenness_bottom}
\small
\setlength{\tabcolsep}{4pt}
\begin{tabular}{llp{7cm}}
\toprule
\textbf{Host} & \textbf{Score} & \textbf{Representative Snippet} \\
\midrule
\texttt{hammarsdrama.com}              & $4.41 \times 10^{-20}$ & Hammars Drama Productions AB. We are Stockholm-based executive producers of performing art and of dance movement films. Board of Directors: Ingmar Bergman jr, Chairman. \\[4pt]
\texttt{easternfirst.applicantpro.com} & $4.09 \times 10^{-20}$ & Eastern Industrial Supplies, Inc. 25-Jun-2018 to 24-Aug-2018 (EST). Greenville, SC, USA. Full Time. Medical, Dental, Disability, Life, 401k + Employer Match \ldots \\[4pt]
\texttt{swadharma.myshopify.com}       & $4.07 \times 10^{-20}$ & ``Hey Julia, how are you and your belly? So do you know what you're having?'' 10 reasons why I enjoyed receiving my prenatal care from my midwife. \\[4pt]
\texttt{ontoma.com}                    & $3.86 \times 10^{-20}$ & An industry-willed response to the Royal Commission into Misconduct in the Financial Services Industry. New FinTech platform Ontoma set to foster cooperation \ldots \\[4pt]
\texttt{bluepenstrokes.com}            & $3.28 \times 10^{-20}$ & Daily Grind. Sublime requests of my creative mind overturned by demands of a cerebral strife. Shackled to cubicles, paints and brushes, paper and ink \ldots \\[4pt]
\texttt{walkertonkinsmen.ca}           & $2.78 \times 10^{-20}$ & Walkerton Kinsmen Raffle Draw \& Novelty Casino Official Rules, Regulations and Draw Procedures. The following are the rules, regulations and draw procedures \ldots \\[4pt]
\texttt{grindstone.agency}             & $2.74 \times 10^{-20}$ & Harmony Honeybush. Packaging design. Brand Development. Our agency was approached to assist with the Brand Development of this new specialist company \ldots \\[4pt]
\texttt{bmscg.com}                     & $2.09 \times 10^{-20}$ & BMS Commercial and Consulting Group. BMSCG leads you to success. We know the right way. The economic power of Asia at your service. \\[4pt]
\texttt{abcsofsex-ed.org}              & $1.98 \times 10^{-20}$ & Teaching an All-Day Workshop for 20 Teachers and Staff. Starkids Academy and Rescue Center, in Kiambu just outside the Nairobi city limits \ldots \\[4pt]
\texttt{riomardesigns.com}             & $6.82 \times 10^{-21}$ & Don't get sick on your next trip! Get your list. 5 Tips for Healthy Travel. There are many things you can do to avoid getting sick next time you travel. \\
\bottomrule
\end{tabular}
\end{table}

\section{Limitations and future work} 

Our experiments are conducted at 400M and 1B parameter scales with 8B and 28B training tokens respectively, following the DCLM 1b-1x reference setting. The scaling pattern we observe—gains that grow with model size—suggests that further improvements may be achievable at larger scales, but verifying this requires substantially more compute. Our centrality scores are also computed at the host level and inherited by all documents from a given host; a finer-grained page-level graph could capture intra-host structural variation that is currently averaged out. We focused on betweenness and Katz centrality because they capture distinct notions of structural importance (cross-community bridging vs.\ recursive influence), but other graph-theoretic measures—including hierarchical decomposition (k-core, k-truss), random-walk-based methods beyond Katz, and motif-based scores—remain unexplored. Finally, combining WebGraphMix with domain-based methods such as WebOrganizer is a natural next step: graph centrality and semantic taxonomies operate on independent axes, and combining them may yield further compounding gains in the same way that combining centrality with content-based~quality~does.

\section{Broader Impact}

This work introduces a graph-based framework for pretraining data selection that operates on the structural topology of the web rather than on document content. We discuss both potential positive and negative societal implications.

\paragraph{Positive impacts.}
WebGraphMix offers a computationally lightweight alternative to data selection methods that require training auxiliary models, running proxy evaluations, or constructing domain taxonomies. By replacing these resource-intensive steps with a one-time centrality computation (fewer than 9 GPU-hours total), our approach lowers the barrier to principled data curation, particularly for resource-constrained research groups. More broadly, improving the efficiency of pretraining data selection reduces the total compute---and therefore energy---spent on training language models, since better data can substitute for additional training steps or larger model sizes.

\paragraph{Potential risks and limitations.}
Graph-based selection introduces a new axis of bias that differs from content-based filtering. Web graph centrality reflects the \emph{linking behavior} of web publishers, which is shaped by commercial incentives, language demographics, and historical web development patterns. Structurally central hosts tend to be large, English-dominant platforms (e.g., social media sites, major reference sites), while peripheral hosts include small organizations, non-English content, and niche communities. Selecting data based on centrality scores therefore risks amplifying the structural inequalities already present in the web's link topology---for example, systematically underrepresenting content from regions or languages with less interconnected web infrastructure.

Our mixture-based approach partially mitigates this concern by explicitly including peripheral documents alongside central ones, and our results show that peripheral regions contribute valuable capabilities that central regions lack. However, we do not conduct a systematic analysis of how centrality-based selection affects demographic, linguistic, or geographic representation in the resulting training data, and we encourage future work in this direction.

Finally, the centrality scores and selection scripts we plan to release are metadata annotations on an already-public corpus and do not introduce new privacy risks beyond those inherent in Common Crawl itself. The models trained in this work are small-scale research artifacts (up to 1.4B parameters) not intended for deployment.


\end{document}